\documentclass{article}

\usepackage{arxiv}

\usepackage[utf8]{inputenc} 
\usepackage[T1]{fontenc}    
\usepackage{hyperref}       
\usepackage{url}            
\usepackage{booktabs}       
\usepackage{amsfonts}       
\usepackage{nicefrac}       
\usepackage{microtype}      
\usepackage{lipsum}		
\usepackage{graphicx}
\usepackage[numbers]{natbib}
\usepackage{doi}
\usepackage{romannum}
\usepackage{color}
\hypersetup{
    colorlinks=true,
    linkcolor=blue,
    citecolor=blue,      
    urlcolor=blue,
}
\usepackage{paralist}
\usepackage{amsmath}
\usepackage{multirow}
\usepackage[table,xcdraw]{xcolor}
\usepackage{longtable}
\usepackage{lscape}
\usepackage{subfig}
\usepackage{graphicx}

\title{Meta-learning approaches for few-shot learning:\\ A survey of recent advances}

\author{ \href{https://orcid.org/0000-0001-8298-7512}{\includegraphics[scale=0.06]{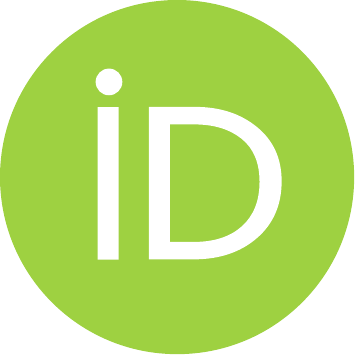}\hspace{1mm}Hassan Gharoun}\\
	Faculty of Engineering \& IT\\
        University of Technology Sydney\\
	NSW, AU\\
	\texttt{Hassan.Gharoun@Student.UTS.edu.au} \\
	\And
	\href{https://orcid.org/0000-0001-5798-917X}           
        {\includegraphics[scale=0.06]{orcid.pdf}\hspace{1mm}Fereshteh Momenifar} \\
	School of Business\\
	Western Sydney University\\
	NSW, AU \\
	\texttt{22046851@student.westernsydney.edu.au} \\
	\AND
        \href{https://orcid.org/0000-0003-4971-8729}{\includegraphics[scale=0.06]{orcid.pdf}\hspace{1mm}Fang Chen} \\
	Faculty of Engineering \& IT\\
        University of Technology Sydney\\
	NSW, AU\\
	\texttt{Fang.Chen@uts.edu.au} \\
 	\AND
        \href{https://orcid.org/0000-0002-2798-0104}{\includegraphics[scale=0.06]{orcid.pdf}\hspace{1mm}Amir H. Gandomi} \\
	Faculty of Engineering \& IT, University of Technology Sydney, NSW, AU\\
        University Research and Innovation Center (EKIK), Óbuda University, Budapest, HU\\
	\texttt{Gandomi@uts.edu.au} \\
}

\renewcommand{\shorttitle}{\textit{Meta-learning approaches for few-shot learning: A survey of recent advances}}

\begin{document}
\maketitle

\begin{abstract}
Despite its astounding success in learning deeper multi-dimensional data, the performance of deep learning declines on new unseen tasks mainly due to its focus on same-distribution prediction. Moreover, deep learning is notorious for poor generalization from few samples. Meta-learning is a promising approach that addresses these issues by adapting to new tasks with few-shot datasets. This survey first briefly introduces meta-learning and then investigates state-of-the-art meta-learning methods and recent advances in: (\romannum{1}) metric-based, (\romannum{2}) memory-based, (\romannum{3}), and learning-based methods. Finally, current challenges and insights for future researches are discussed. 
\end{abstract}


\section{Introduction}

\par Humans possess the extraordinary capability of learning a new concept even after minimal observation. To a greater extent, a child can distinguish a dog from a cat through a single picture \cite{vinyals2016matching}. This critical characteristic of human intelligence lies in the humans’ ability to leverage obtained knowledge of prior experiences to unforeseen circumstances with small observation. \par
Unlike the human learning paradigm, traditional machine learning (ML) and deep learning (DL) models train a specific task from scratch through: (a) the training phase in which a model is initiated randomly and then updated, and (b) the test phase in which the model evaluates. While ML and DL have obtained remarkable success in a wide range of applications, they are notorious for requiring a huge number of samples to generalize. In many real-world problems, collecting more data is costly, time-consuming, and even might not feasible due to physical system constraints \cite{lu2020learning}. Moreover, most ML and DL models presume that training and testing datasets have the same distribution \cite{wang2022generalizing}. Thus, their performance suffers under data distribution shifts \cite{wang2020tent}. While the model's generalization ability significantly depends on data distribution, training a model on a diverse dataset does not result in domain adaptation \cite{finn2018learning}. Domain adaptations aim to transfer obtained knowledge from prior experiences to learn new but relevant tasks \cite{wang2018deep}. Transfer learning, a well-established sub-field in machine learning, uses knowledge of related source domains with sufficient annotated data to improve the model's performance on related target domains with inadequate or lack of labeled data \cite{farahani2021brief}. Model parameters, feature representations, and instances are different forms of knowledge that can be transferred in transfer learning \cite{pan2009survey}. In addition, transfer learning has achieved tremendous success in a wide range of applications, such as computer vision \cite{shaha2018transfer} and natural language processing \cite{ruder2019transfer}. Arguably, transfer learning has obtained outstanding achievement for supervised pre-training of convolutional neural networks (CNNs) on ImageNet. However, the performance of the pre-trained network deteriorates during fine-tuning with few samples \cite{ravi2016optimization}, or new samples become less similar \cite{yosinski2014transferable}. One way to generalize unseen samples directly from few examples is by learning rich knowledge from similar events and re-using past knowledge to incrementally adapt to unseen examples. This idea, motivated by humans' intelligence, is known as meta-learning. \par
\begin{figure}[!h]
    \centering
    \includegraphics[width=0.7\linewidth]{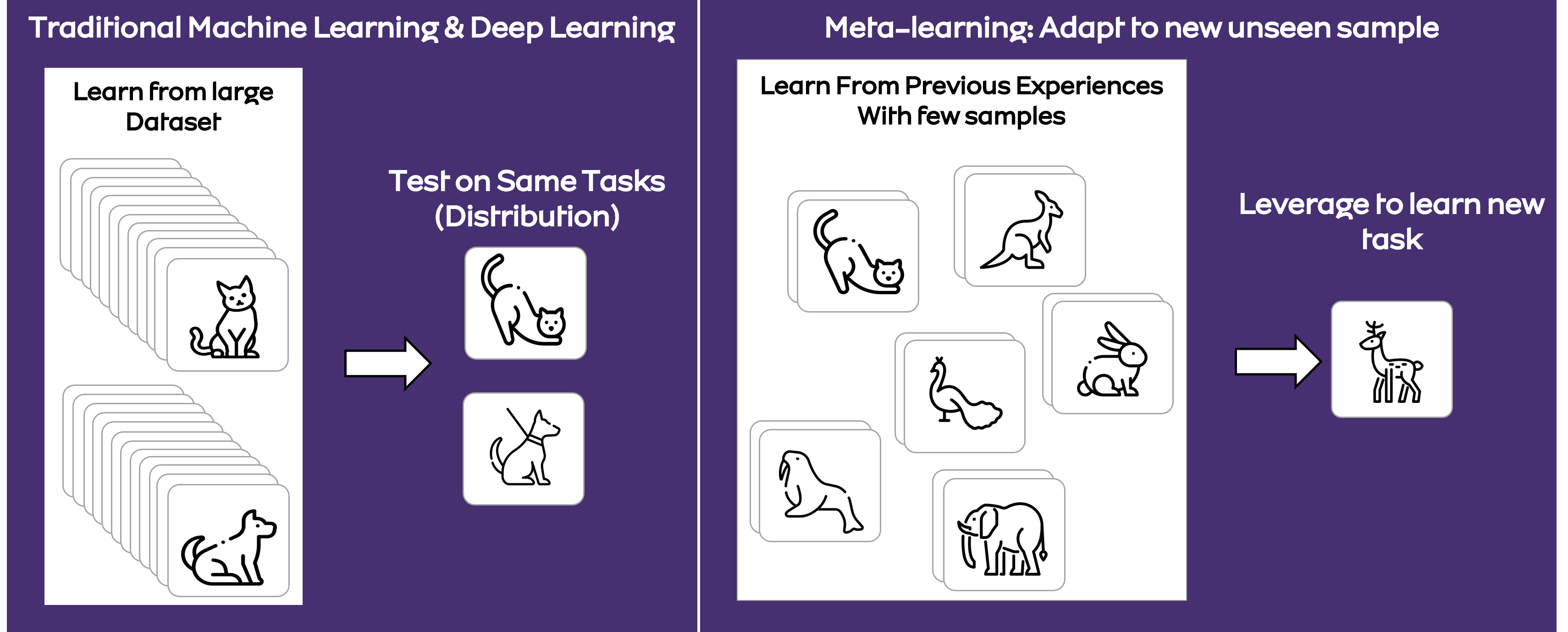}
    \caption{Standard Machine learning and deep learning vs. meta-learning paradigm}
    \label{fig_learning_to_learn}
\end{figure}

Meta-learning refers to a variety of techniques that aim to improve adaptation through transferring generic and accumulated knowledge (meta-data) from prior experiences with few data points to adapt to new tasks quickly without requiring training from scratch. Meta-learning is most commonly known as \textit{”Learning to learn”}. But, what does learning to learn mean? Mitchel \cite{mitchell1997machine} defined the term \textit{learning} as the instance when a computer program's performance at a given task improves with experience. In conjunction with Mitchel's definition, Thrun and Pratt \cite{thrun2012learning} defined \textit{”Learning to learn”} as a capability of algorithms to improve their tasks’ performance with experience and with a number of observed tasks. In this context, an algorithm's performance improvement depends on a family of tasks. In other words, if a learning algorithm's performance does not improve with the experience of other tasks, it is not said to learn to learn \cite{thrun2012learning}. Accordingly, the learning level is elevated from data to tasks in meta-learning. Figure \ref{fig_learning_to_learn} better demonstrates the learning to learn paradigm.
Driven by the goal to scale AI to a real-world setting and approaches to the human intelligence paradigm, meta-learning has recently regained significant interest in few-shot problems using deep learning \cite{wang2021meta}. Currently, a number of surveys on meta-learning have been proposed \cite{huisman2021survey, li2021concise}, however, the meta-learning methods progress rapidly. Accordingly,
this paper examines meta-learning methods in detail and performs an extensive review of meta-learning to provide an up-to-date review. Additionally, a new organization from the algorithms' mechanics point of view is introduced. A summary of the benchmark comparison of review methods is also presented, followed by a comprehensive discussion of challenges to establish future directions based on current limitations and strengths.

The remainder of this paper is organized as follows. Section \ref{section_Preliminary} reviews the preliminary concepts of meta-learning. Section \ref{section_Meta-Learning_review} introduces the category of meta-learning methodologies and presents an in-depth review of state-of-the-art methods and their advances. Finally, section \ref{section_challenges} describes the challenges along with future research directions and brief conclusions.

\section{Meta-Learning Preliminary Terminology and Training Structure}  \label{section_Preliminary}

\subsection{Terminology and Training Structure}
In the standard supervised machine learning setting, dataset $D$ is divided into two separable and distinct datasets, namely train and test sets denoted by $D_{train}$ and $D_{test}$, respectively. Conventional neural networks train during iterative cycles, termed \textit{"Epoch"}. In every epoch, the entire $D_{train}$ is feeding forward and backward. During each epoch, $D_{train}$ may be partitioned into several packets termed \textit{"Batch"}. Thus, each epoch may consist one or several \textit{"iterations"}, which is the number of steps needed for one epoch that all batches (i.e., entire $D_{train}$) feed into the model. \par

\begin{figure}[!h]
    \centering
    \includegraphics[width=\linewidth]{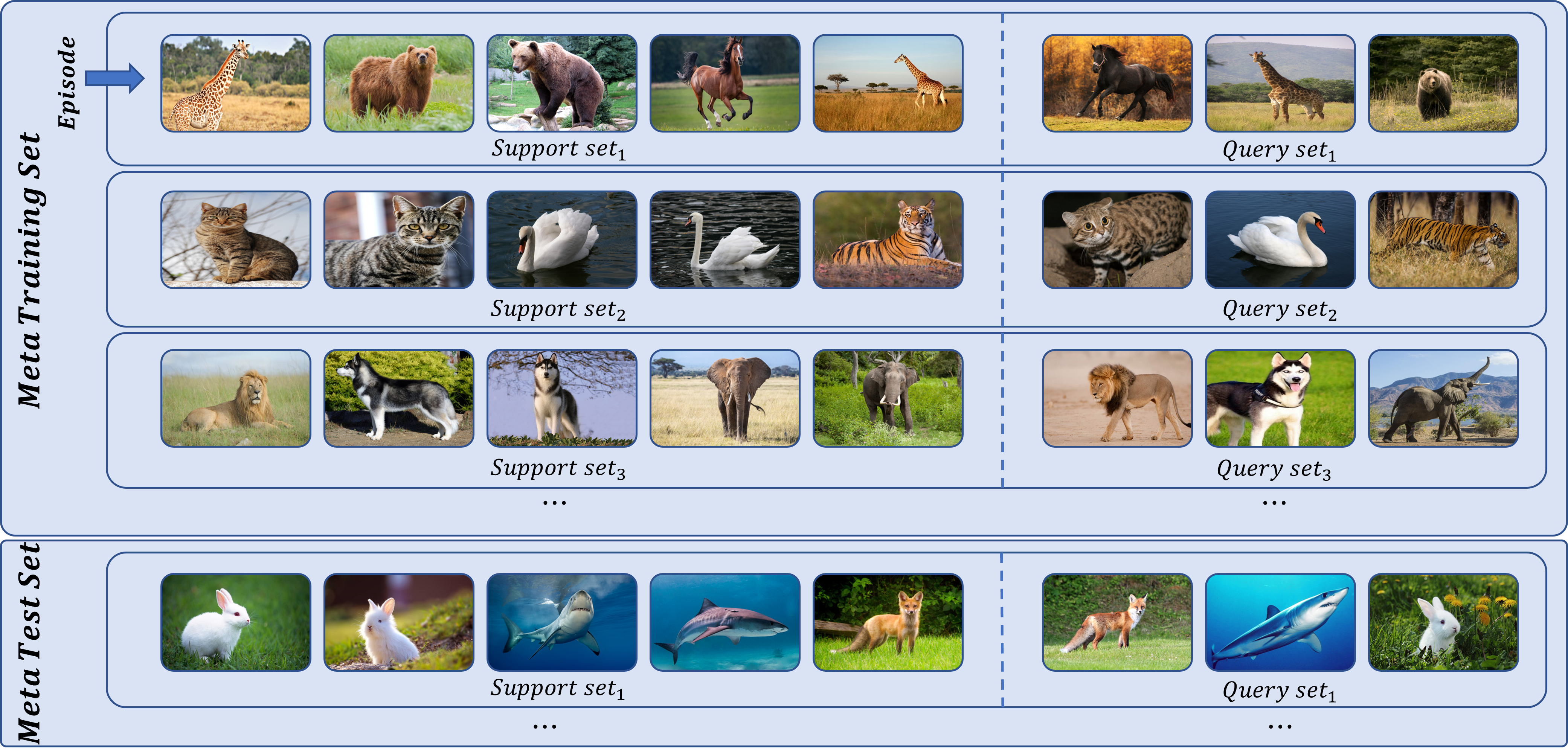}
    \caption{Meta training and test set, Episodic training schema}
    \label{fig_Episodic_Training}
\end{figure}

In meta-learning, dataset $D$ is divided with no overlapping into two meta sets, namely the meta train  and meta test sets, denoted by $D_{meta\_train}$ and $D_{meta\_test}$, respectively. Next, $D_{meta\_train}$ and $D_{meta\_test}$ are subdivided into train and test sets. To differentiate between these two forms of training and testing sets, the latter (inner) train and test sets are termed \textit{"Support set"} and  \textit{"Query set"}, respectively. Let's rewrite the meta train set and meta test set as $D_{meta\_train}=\left \{ (D_{meta\_train}^{Support}, D_{meta\_train}^{Query}) \right \}$ and $D_{meta\_test}=\left \{ (D_{meta\_test}^{Support}, D_{meta\_test}^{Query}) \right \}$. Support set is a \textit{N-way k-shot} random samples and the query set is q random samples for each of the $N$ categories of the support set. A meta-learner model trains in an \textit{"Episodic"} mode. In each episode, support and query sets would created. A meta-learner model takes $D_{meta\_train}^{Support}$ to learn, then applies on the $D_{meta\_train}^{Query}$ to make predictions. The meta-learner updates based on the prediction error over episodes. Over a series of episodes, the meta-learner learns to learn from the small dataset. This phase is called meta-learning. 
Afterward, it takes $D_{meta\_test}^{Support}$ to build a classifier. Here, the model concentrates on learning task-specific parameters quickly. Then, the performance of the classifier is evaluated through $D_{meta\_test}^{Query}$, which is known as the adaptation phase. In other words, the model uses the prior knowledge learned in the meta-learning phase to quickly adapt to a new task in the adaptation phase. From a conventional neural networks view, episodic training corresponds to training on a series of batches \citep{laenen2021episodes}. A visualization of data preparation and episodic training is demonstrated in Figure \ref{fig_Episodic_Training}.

\subsection{Datasets}
Most state-of-the-art meta-learning models have been evaluated and compared on several well-known benchmarks. While there several datasets exist, the three most widely used benchmark datasets are:
\vspace{-\topsep}
\begin{itemize}
    \item Omniglot: an image dataset of different handwritten characters of different alphabets \citep{lake2011one}. This dataset is comprised of 1623 characters related to 50 different alphabets. In the few-shot learning setting, each character stands as classes/labels, and each character has 20 samples (or 20 shots) drawn by 20 different person. 
    \item miniImageNet: the mini version of the popular large ImageNet dataset published in 2009, which revolutionized the artificial intelligence realm. ImageNet is a large colored image dataset annotated by humans for visual recognition competition. Now, ImageNet contains 14 million images and covers 20,000 objects/classes. For few-shot learning problems, \cite{vinyals2016matching} published miniImageNet from ImageNet, a smaller dataset containing 100 classes with over 60,000 images. 
    \item CUB-200-2011: the Caltech-USCD-Birds-200-2011 \citep{wah2011caltech}, in short CUB-200-2011, contains 11,788 images of 200 different bird species (200 classes). Each image has a textual description alongside labels describing various attributes, such as color and shape. 
\end{itemize}
Oxford-102 (flowers dataset), CIFAR-10, and MNIST are other datasets that are less used in the literature. In this paper, the above three datasets are used to compare the performance of different methods. 
\section{Meta-Learning} \label{section_Meta-Learning_review}

\begin{figure}[!h]
    \centering
    \includegraphics[width=8cm]{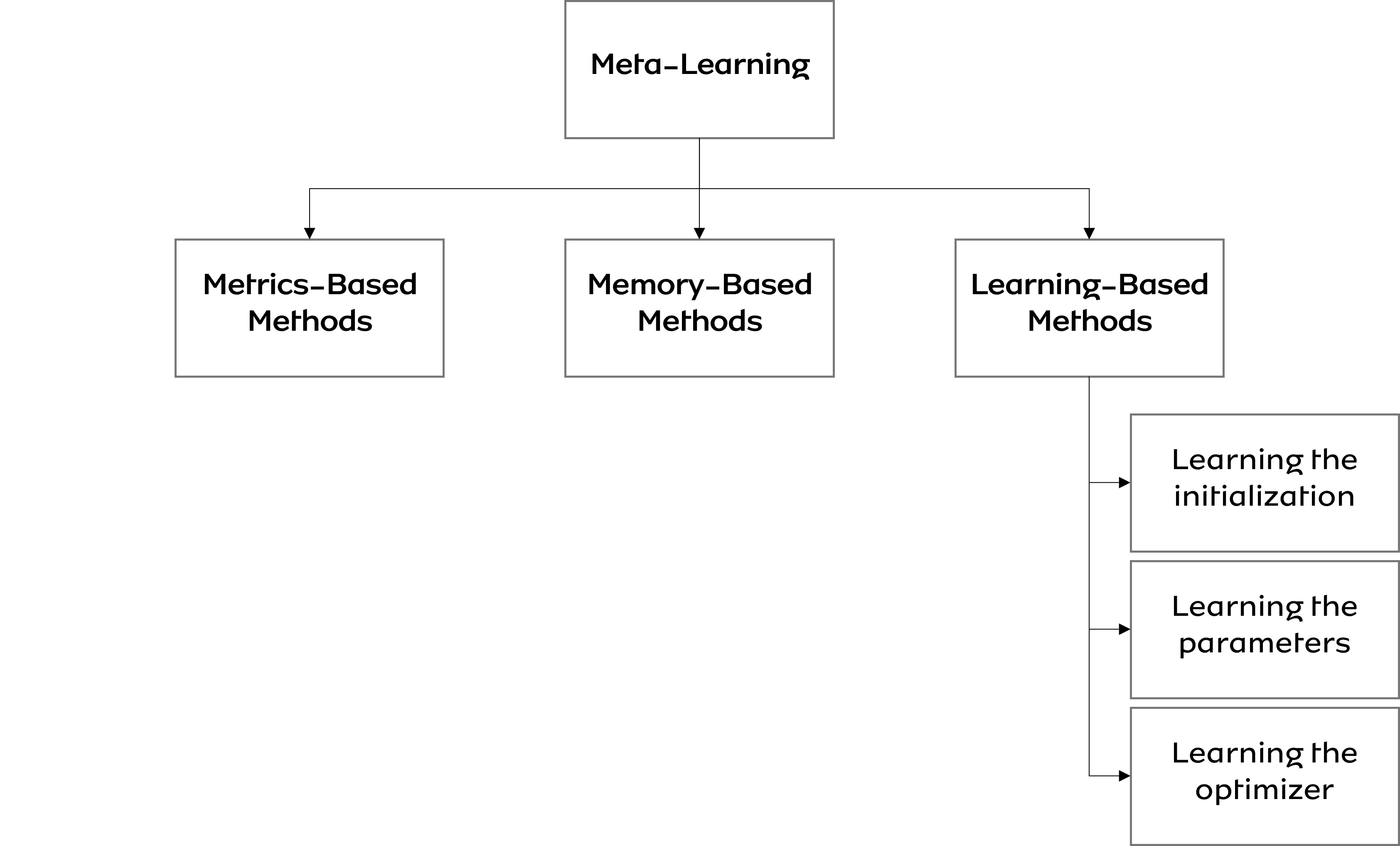}
    \caption{Meta-Learning methods categories}
    \label{fig_meta_categories}
\end{figure}

This section provides a comprehensive overview of cutting-edge studies and their advances in meta-learning in three main categories (shown in Figure \ref{fig_meta_categories}): 
\begin{compactitem}
    \item Metric-based methods
    \item Memory-based methods
    \item learning-based methods
    \begin{itemize}
        \item Learning the initialization
        \item Learning the parameters
        \item Learning the optimizer
    \end{itemize}
\end{compactitem}

\subsection{Metric-based methods}
The key concept in "learning the metric space" methods is comparing the query set samples to support set samples. Then, the labels of query samples are predicted based on proximity to the support set. Principally, the success of these methods depends on embedding learning. In this context, the aim of embedding learning is to convert each high dimensional sample to a lower dimensional sample in the form of vector representations in such a way that similar samples reside closely in a new lower dimension space. Given an N-way k-shot problem, let's consider a support set $D_{support} = \left \{ (x_{1}^{s}, y_{1}^{s}), \cdots, (x_{K}^{s}, y_{K}^{s}) \right \}$, where $x_{i}^{s}$ refers to input samples, $y_{i}^{s}\in N$ is predefined labels, and a query set $D_{query} = \left \{ (x_{1}^{q}, y_{1}^{q}), \cdots , (x_{J}^{q}, y_{J}^{q}) \right \}$, in which $x_{j}^{q}$ is the test sample and $y_{j}^{q}\in N$ is the true label of test. In a generic form of learning the metric space methods, embedding function $f$ and embedding function $g$ apply on $x_{i}^{s}$ and $x_{j}^{q}$, respectively, to generate embedding of the support set and query set. Next, the similarity between obtained embeddings, i.e., $f(x_{i}^{s})$ and $g(x_{j}^{q})$, is measured by similarity function $s(\cdot,\cdot)$. To predict the label of $x_{j}^{q}$, a decision is made based on the closest similarity between $f(x_{i}^{s})$ and $g(x_{j}^{q})$. In other words, the class $f(x_{i}^{s})$ nearest to to $g(x_{j}^{q})$ is assigned to the sample test $x_{j}^{q}$. While the same embedding function can be used for both support and query sets, several researches have shown that different embedding functions lead to a better performance \citep{vinyals2016matching, bertinetto2016learning}. Current state-of-the-art metric-based algorithms are:\par

\subsubsection{Siamese Network}\par

\begin{figure}[!h]
    \centering
    \includegraphics[width=8cm]{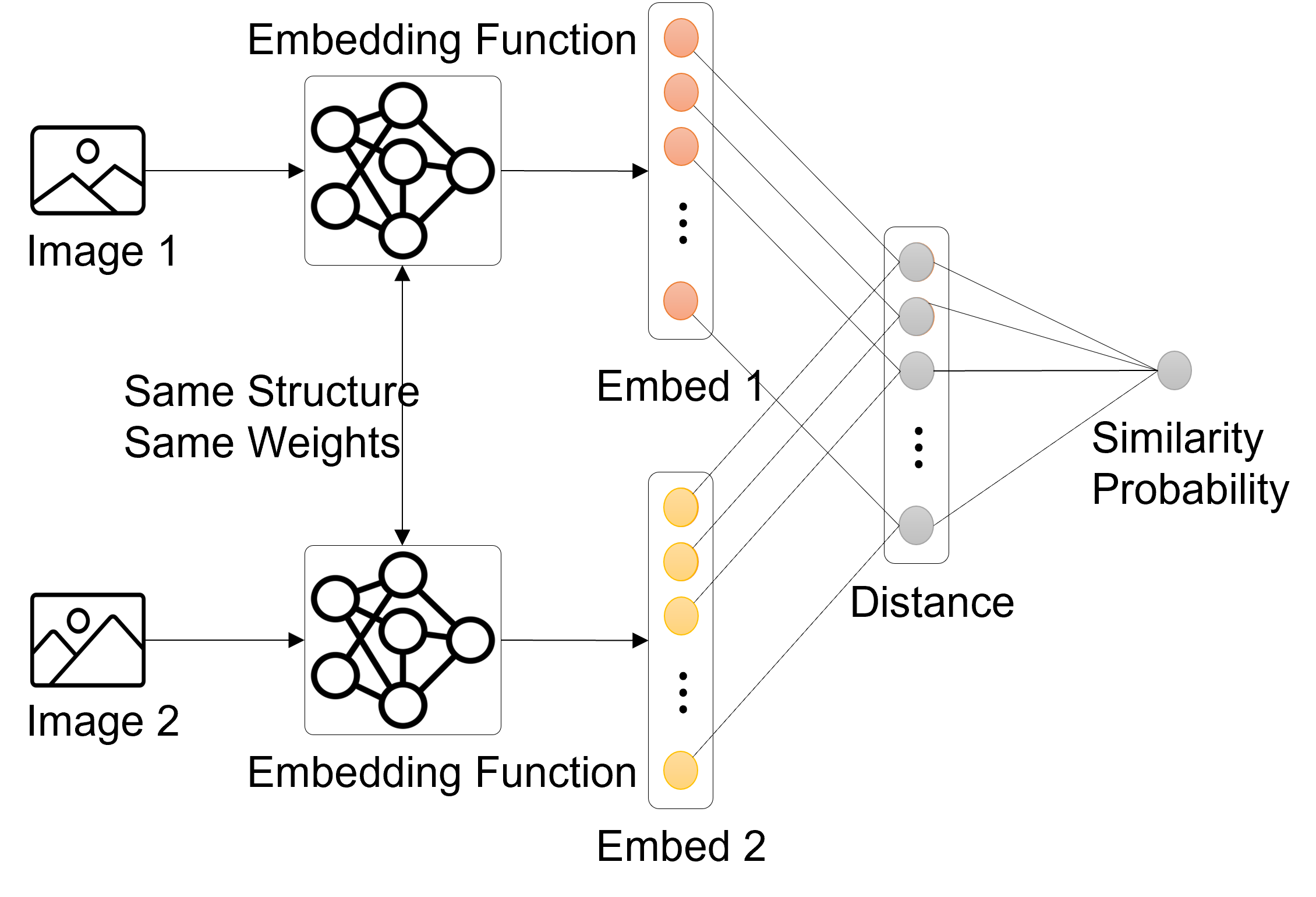}
    \caption{Siamese architecture for one-shot classification problem}
    \label{fig_siames_architecture}
\end{figure}

Siamese network is one of the simplest metric-based learners widely used in one-shot learning problems. The principal architecture of the Siamese network was proposed in 1993 by \cite{baldi1993neural} for fingerprint similarity estimation. However, the term "Siamese" was first introduced by \cite{bromley1993signature} for the signature verification problem. Principally, the Siamese network consists of two symmetrical neural networks joined with the similarity function called the "energy function." The architecture of the Siamese network is shown by Figure \ref{fig_siames_architecture}. 

As demonstrated in Figure \ref{fig_siames_architecture}, a input pair $x_1$ and $x_2$ from the training set is fed to two embedding functions, $f_1(\cdot)$ and $f_2(\cdot)$, sharing the same parameters. Then, embeddings $f_1(x_1)$ and $f_2(x_2)$ feed into energy function. Basically, the energy function could be any similarity measure, such as cosine similarity or Euclidean distance. The output of the energy function is the similarity score of the two embeddings. Thus, rather than a classification task, the Siamese network principally aims to learn the network for matching tasks between similar pairs \citep{melekhov2016siamese}. The proposed network utilizes contrastive loss as a loss function for optimization. The traditional contrastive loss used in the generic Siamese network is expressed as Eq.\ref{eq_traditional_contrast_loss} \citep{chopra2005learning, ilina2022survey}:

\begin{equation}
    \mathfrak{L}(E,Y) = \sum_{i=1}^{P} (1-Y).L_G(E(Z_1, Z_2)_i) + Y.L_I(E(Z_1, Z_2)_i)
    \label{eq_traditional_contrast_loss}
\end{equation}

\noindent where $E(X_1, X_2)^i$ is the energy function that measures the similarity of $i^{th}$ pairs of embeddings $X_1$ and $X_2$. Y is the true label that denotes whether the given pair is similar $Y=0$ or dissimilar $Y=1$. $L_G$ and $L_I$ are partial loss functions applied on similar and dissimilar pairs, respectively, and designed in a way that monotonically increases the loss function for a similar pair and decreases loss function for a dissimilar pair. Siamese networks are widely used in various applications where data are limited. The face verification task is one of the extensive applications of the Siamese network. In the state-of-the-art research, \cite{chopra2005learning} studied the task of face verification by refining the loss function as Eq. \ref{eq_max_margin_contrastLoss}:

\begin{equation}
    \mathfrak{L}(E,Y) = (1-Y).E^2 + Y.max(0, margin - E)^2
    \label{eq_max_margin_contrastLoss}
\end{equation}

\noindent where the margin is a hyper-parameter that specifies the lower bound distance between dissimilar pairs. Similar to Eq.\ref{eq_traditional_contrast_loss}, $E$ is the energy function, and $Y$ will be 0 for the similar pair and 1 otherwise. This study was one of the earliest works that used convolutional neural networks (CNNs) as an embedding function. Later, \cite{koch2015siamese} studied Siamese networks for a one-shot image recognition problem, which  is known as the first study of a metric-based method extended Siamese networks for few-shot learning. Additionally, Siamese networks have been extensively studied in various applications. For further details on Siamese network applications, interested readers can refer to comprehensive reviews by \cite{chicco2021siamese, ilina2022survey}.

\subsubsection{Prototypical Networks}\par

\begin{figure}[!h]
    \centering
    \includegraphics[width=0.65\linewidth]{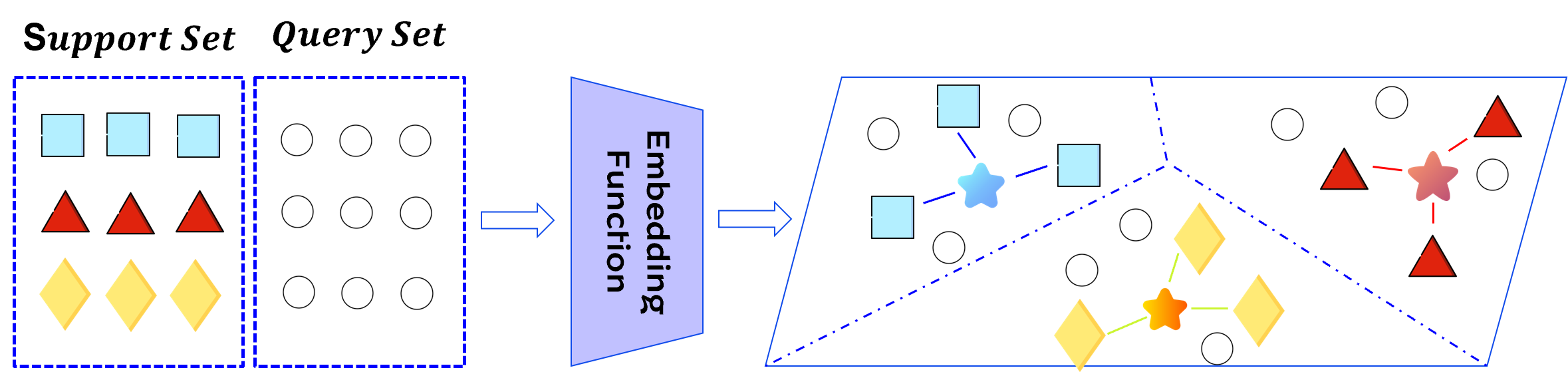}
    \caption{Architecture of Prototypical Networks.}
    \label{fig_prototypical_architecture}
\end{figure}

The prototypical network, proposed by \cite{snell2017prototypical}, is basically learning an embedding function through clustering the embedding space. 
Principally, prototypical networks aim to generate a prototype of each class to use as an anchor for the classification task. Given the N-way k-task classification problem, the prototypical representation of each class is built based on the mean embedding of $k$ data points in each N class. By using the same embedding function for both the support and query sets, the embedding of query points would obtained. Mainly, Euclidean distance is used as a distance metric, measuring the distance of query embeddings and classes' prototypes. Each query point is classified based on the closest distance of the query embedding and classes' prototypes. Predictions probabilities are obtained by feeding query-prototypes distance into the Soft-max function. The prototypical network trains by minimizing the average negative log-probability of the correct prediction, as expressed by Eq.\ref{eq_prototypical_nll}. Figure \ref{fig_prototypical_architecture} shows the general flow of the prototypical networks. 

\begin{equation}
    -\frac{1}{T}\sum_{i}log p(y_i^*|x_i^*,{p_c})
    \label{eq_prototypical_nll}
\end{equation}

Two well-known variants of prototypical networks are semi-supervised and Gaussian prototypical networks. \cite{ren2018meta} revised the prototypical network to deal with the semi-supervised setting, where the dataset contains both labeled and unlabeled data. In their model, similar to the generic prototypical network, class prototypes are first built with labeled data. Then, soft k-means algorithm is applied on the embeddings of both labeled and unlabeled data. However, inspired by seeded k-means, instead of initializing cluster centers randomly, the class prototypes are calculated from the labeled data, and are used as cluster centers. Then, unlabeled embeddings are assigned to each cluster based on their Euclidean distance to each cluster center, i.e., class prototypes. Finally, class prototypicals are refined by updating cluster centers by the usual k-means algorithm. Since some points that do not belong to any existing classes may exist in unlabeled data, these data would interfere with prototypes refinements. Thus, the authors proposed an additional class for these data called the distractor class and considered the origin as the center of the distractor class. Since distractor class members may belong to different subjects, a single distractor class will have a high variance. To solve this issue, the authors proposed a soft-masking mechanism, in which a threshold distance is set for all original class prototypicals. This threshold acts as an intra-cluster variance capturing distractors as exemplars outside the prototypical areas. This threshold is calculated by feeding statistics of prototypes, including min, max, skewness, and kurtosis, to a small multi-layer perceptron (MLP) network.\par
\begin{figure}[!h]
    \centering
    \includegraphics[width=0.7\linewidth]{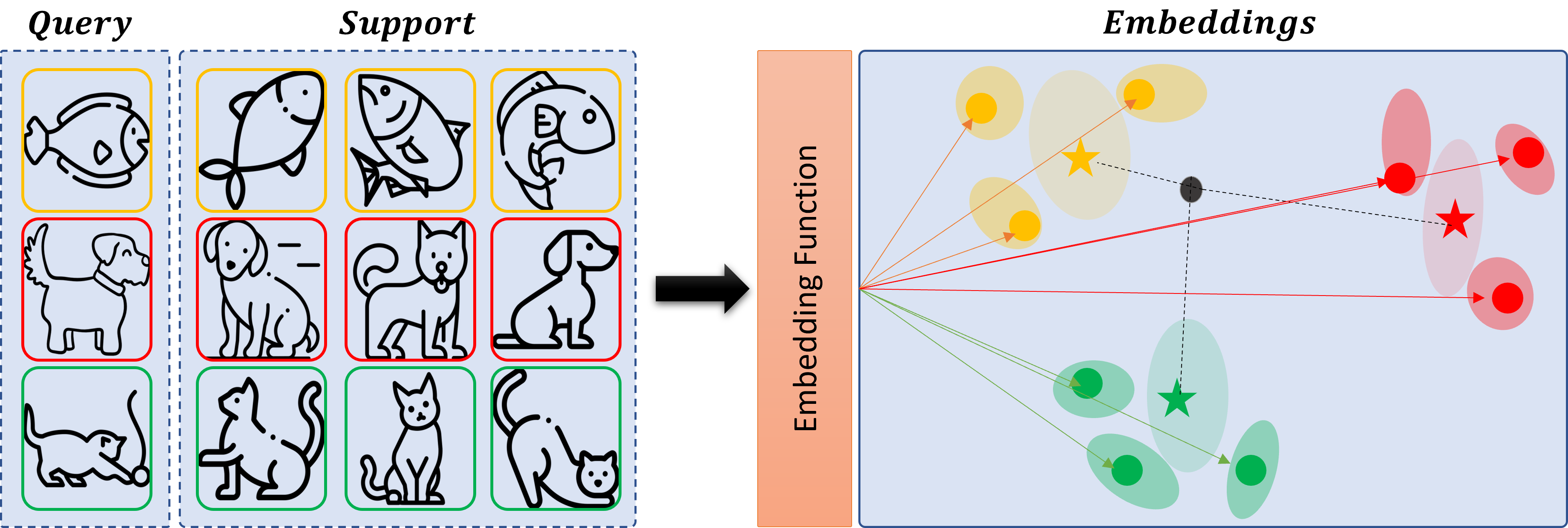}
    \caption{Gaussian prototypical networks workflow; Solid circles are embedding vectors, dark ellipses represent each embedding covariance matrix,  and light halos depict the class's total covariance matrix. Finally, stars illustrate classes' prototypicals.}
    \label{fig_gaussian_PN}
\end{figure}

The Gaussian prototypical network is another novel architecture of the prototypical network proposed by \cite{fort2017gaussian}. In this model, a covariance matrix is constructed from the embedding function alongside the generated embedding vectors. 
The covariance matrix is used to build a confidence region around data points and adds weights to embedding vectors, then utilized to compute a total covariance matrix of each class. Afterward, the distance of each query from class prototypes is used to compute the linear Euclidean distance as follows (Eq.\ref{eq_gaussian_Euclid_dist}):

\begin{equation}
    d_c^2(i) = (\vec{x_i}-\vec{p_c})^TS_c(\vec{x_i}-\vec{p_c})
    \label{eq_gaussian_Euclid_dist}
\end{equation}

\noindent where $p_c$ is the cluster center or the prototype of class $c$; $S_c$ is the inverse covariance matrix; $d_c$ is the distance of query point $i$ from class prototype $c$; and $\vec{x_i}$ is query point $i$. Figure \ref{fig_gaussian_PN} illustrates the operation of Gaussian prototypical networks. Variable $p_c$  is calculated using variance-weighted embedding vectors in Eq.\ref{eq_gaussian_cov_matrix}:

\begin{equation}
    \vec{p_c} = \frac{\sum_{i}\vec{s_i}^c.\vec{x_i}^c}{\sum_{i}\vec{s_i}^c}
    \label{eq_gaussian_cov_matrix}
\end{equation}

\noindent where $\vec{x_i}^c$ is the embedding vector of support point $i$ and class $c$; and $\vec{s_i}^c$ is the covariance matrix of related support point $i$ and class $c$. Class covariance matrix is computed as Eq.\ref{eq_gaussian_class_cov}:

\begin{equation}
    \vec{S_C} = \sum_{i}\vec{s_i}^c
    \label{eq_gaussian_class_cov}
\end{equation}

 \cite{pan2019transferrable} proposed the transferable prototypical network (TPN) by remoulding prototypical networks to learn unsupervised domain adaptation. TPN uses labeled data as a source and unlabeled data as target domains and tries to learn a neural network that minimizes embedding shifts between source and target domains. In other words, TPN constructs an embedding function that generates domain-invariant representations. Technically, TPN first builds class prototypes from the source domain (i.e., labeled data), then the target domain (i.e., unlabeled data) assigned to the nearest class prototypes. The authors proposed three different classifiers regarding three datasets, namely source only data, target only data, and combined source and target data. The authors argued that if the source and target domain distributions are the same, the class prototypes on both domains should be identical. Thus, to achieve the representation invariant over domains, they compute class level discrepancy over domains by calculating reproducing kernel Hilbert space (RKHS). Finally, TPN trains by minimizing combined classification loss and discrepancy loss. 

With a similar belief and similar mechanics to the transferable prototypical network (TPN) \citep{pan2019transferrable}, \cite{lu2018boosting} introduced a new module called domain alignment component to prototypical networks to tackle the problem of domain shifts between test and train sets. Basically, the domain alignment component tries to minimize the maximum mean discrepancy (MMD) by calculating the similarity of two different domains, one from the train set and one from the test set, in kernel Hilbert space. Then, the networks are trained to minimize the combined negative log-probability loss (the generic prototypical networks loss) and MMD measure. Empirical analysis on benchmark datasets showed that their model outperforms the generic prototypical networks. 

\cite{ji2020improved} argued that most metric-based meta-learning models forget intra-class distribution information. To tackle this problem, they proposed another tweak to prototypical networks called improved prototypical network (IPN), which has three main components: (\romannum{1}) feature extraction, (\romannum{2}) weight distribution, and (\romannum{3}) distance scaling. The first component basically extracts feature embeddings and constructs class prototypes. In generic prototypical networks, samples of the same class have the same importance (weight) in class prototypes. The authors refined this idea by proposing an attention-analogous strategy. Hence, the second component, weight distribution, is responsible for computing weighted class prototypes. The weight distribution component is designed as a three-layer neural network that takes concatenated embeddings of the same class as the input and provides samples' weights as the output. The last component, distance scaling is a three-layer neural network that maximizes inter-class differences and minimizes the intra-class differences. The results indicated that IPN outperforms the generic prototypical networks. 

\cite{liu2022label} expanded the prototypical network to multi-label few-shot problems. The authors argued that the main challenge with multi-label problems is that different labels can have the same support set leading to the same prototypes. To address this issue, they proposed a new prototypical network called label-enhanced prototypical network (LPN). For a multi-label few-shot problems, the authors targeted the aspect category detection in sentiment analysis. Accordingly, their data points and class labels consisted of sentences and aspect categories. The LPN is comprised of four main modules: (\romannum{1}) feature extraction, (\romannum{2}) label-enhanced prototypical network, (\romannum{3}) contrastive learning, and (\romannum{4}) adaptive multi-label inference. The authors used a pre-trained embedding function for language, such as Bert \citep{devlin2018bert} as the feature extraction module. The second module is responsible for computing the relation between data points, here sentences, and labels. For this purpose, after extracting support sentence representations from the previous module, the same process is repeated for labels that are verbal phrases to get label representations. The authors used the low-rank bilinear model \citep{yu2018beyond} to assign weight to each sentence's representations. In simple words, the authors calculated the sentence representations' weights by element-wise multiplication of sentences' representations and their label representations, normalized with the Soft-max function. Then, class prototypes are built based on the weighted sentences' representations. The label-enhanced prototypical network module is trained to minimize the cross-entropy loss. The third module is designed in a way to maximize intra-class similarity and minimize inter-class similarity by employing contrastive learning loss. Since in their problem, every sentence can have a different number of labels as a multi-label problem, the next challenge was determining the number of labels that should predict for each new query point. This challenge was addressed by proposing the fourth module, which is basically a multi-layer perception network that predicts the number of labels for each sentence. This network is trained to minimize the cross-entropy loss of predicted label counts. By integrating all modules and combining all losses, LPN outperforms the generic prototypical network. \cite{deuschel2021multi} proposed another variant to prototypical named the multi-prototypes prototypical network. In contrast to the generic prototypical networks that consider only one prototype per class, the proposed network investigates the idea of multiple prototypes per class. For this purpose, the authors partitioned the embedding space of each class into $k$ clusters by employing the k-means clustering algorithm. Class prototypes are simply cluster centers, and the proposed network is trained to minimize the COREL loss. The results showed that the number of class clusters, i.e., classes prototypes affects model performance; in other words, the difficulty of determining the optimal number of prototypes per class limits the performance.

\cite{kim2022dummy} argued that most metric-learner models are not able to distinguish classes out of support set classes and it is necessary for meta-learner models to be able to detect unseen classes during prediction. The authors proposed dummy prototypical networks (D-ProtoNets) to address this issue. Conceptually, D-ProtoNets follows the same mechanics of the generic prototypical networks, but in every episode, a dummy prototype is built by introducing a dummy generator function. In simple words, the dummy generator is a Maxpool layer followed by non-linear fully connected layers performed on the support classes' prototypes. The dummy prototypes are responsible for handling unseen classes. The authors targeted the keyword spotting detection problem and created and published splitGSC, a benchmark dataset for key spotting detection problems based on the Google speech commands dataset. Through various numerical analyses, the authors demonstrated that D-ProtoNets achieved state-of-the-art performance on splitGSC.

\begin{figure}[!h]
    \centering
    \includegraphics[width=0.5\linewidth]{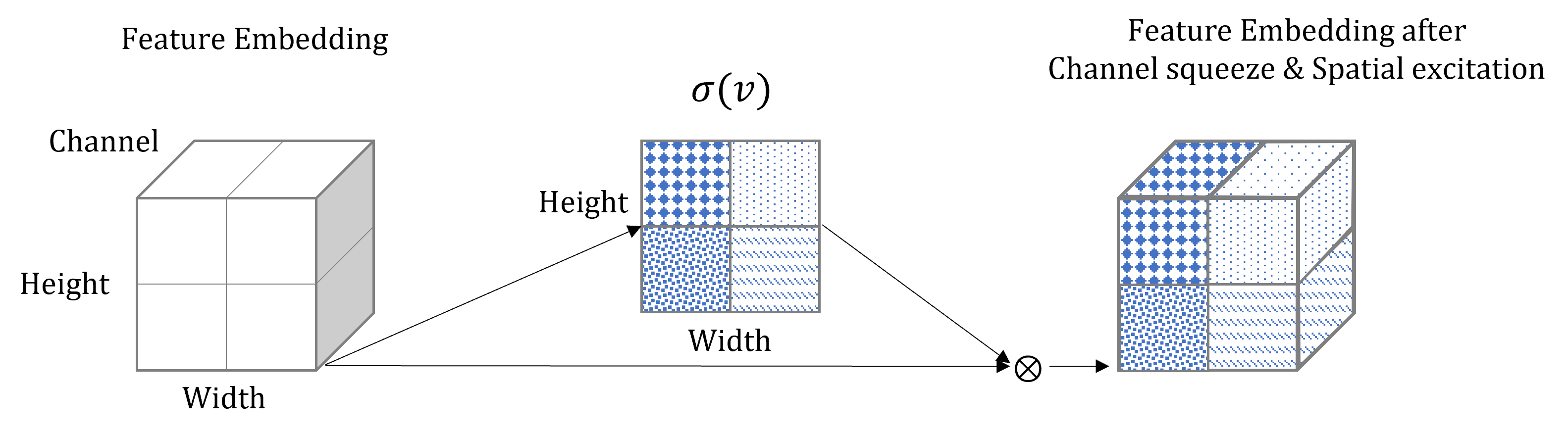}
    \caption{Schema of channel squeeze and spatial excitation components. $\sigma(v)$ corresponds to the relative positional importance of the information of embeddings.}
    \label{fig_Channel_Squeez_Spatial}
\end{figure}

\cite{huang2021local} argued that a single prototype for each class, like the generic prototypical network, is not able to capture the class distribution effectively, so they proposed a local descriptor-based multi-prototype network (LMPNet). Technically, LMPNet utilizes a local descriptor to obtain feature embeddings. The local descriptor is basically the two-dimensional vector output from a convolutional layer \citep{wei2017selective}. LMPNet builds class prototypes based on the weighted average of the local descriptors. These prototypes were obtained through a novel mechanism by introducing channel squeeze and spatial excitation components, which flatten feature embeddings regarding their spatial position over channel dimensions (Figure \ref{fig_Channel_Squeez_Spatial} ). In summary, this attention module is designed to learn the output of multiple prototypes from local descriptors compressed by channel squeeze and spatial excitation. Using Resnet as a backbone, LMPNet outperforms the generic matching network.

\cite{pahde2021multimodal} extended prototypical networks to a multi-modal prototypical network. The core novelty of this model is the use of textual descriptions attached to the image dataset to create more feature embeddings. For this purpose, the authors trained a text-conditional generative adversarial network (tcGAN), which creates image features from the given textual description. Then, two prototypes per class were calculated – one from the original images' embeddings and one from the generated images' embeddings. Next, combined class prototypes are computed based on the weighted average of both original and generated prototypes. Finally, the multi-modal prototypical network was trained in a similar fashion to generic prototypical networks.

\subsubsection{Matching Networks}
Matching networks are another powerful metric-based meta-learner proposed by \cite{vinyals2016matching} that are able to generate labels for the unobserved class as well. In this work, a parametric neural network $P$ was developed to produce a prediction for each query point from the support set $S =\left \{ \left ( x_i,y_i \right ) \right \}_{i=1}^k$, ($k$ number of sample per class), per Eq.\ref{eq_matching_network_p}:

\begin{equation}
    p\left ( \hat{y}|\hat{x}, S \right ) = \sum_{i=1}^{k}a\left ( \hat{x}, x_i \right )y_i
    \label{eq_matching_network_p}
\end{equation}

\noindent where $p\left ( \hat{y}|\hat{x}, S \right )$ is the predicted class for the query point $\hat{x}$ given support set $S$; $x_i$ and $y_i$ are the inputs and labels of samples from support set $S$; and $a$ is the attention mechanism between $\hat{x}$ and $x_i$. The attention mechanism is simply the Soft-max function over the cosine distance between embeddings of $\hat{x}$ and $x_i$, which are calculated via Eq.\ref{eq_matchingNet_attention_mech}: 

\begin{equation}
    a\left ( \hat{x}, x_i \right ) = \frac{e^{cosine(f(\hat{x}),g(x_i))}}{\sum_{j=1}^{k}e^{cosine(f(\hat{x}),g(x_j))}}
    \label{eq_matchingNet_attention_mech}
\end{equation}

\begin{figure}[!h]
    \centering
    \includegraphics[width=0.5\linewidth]{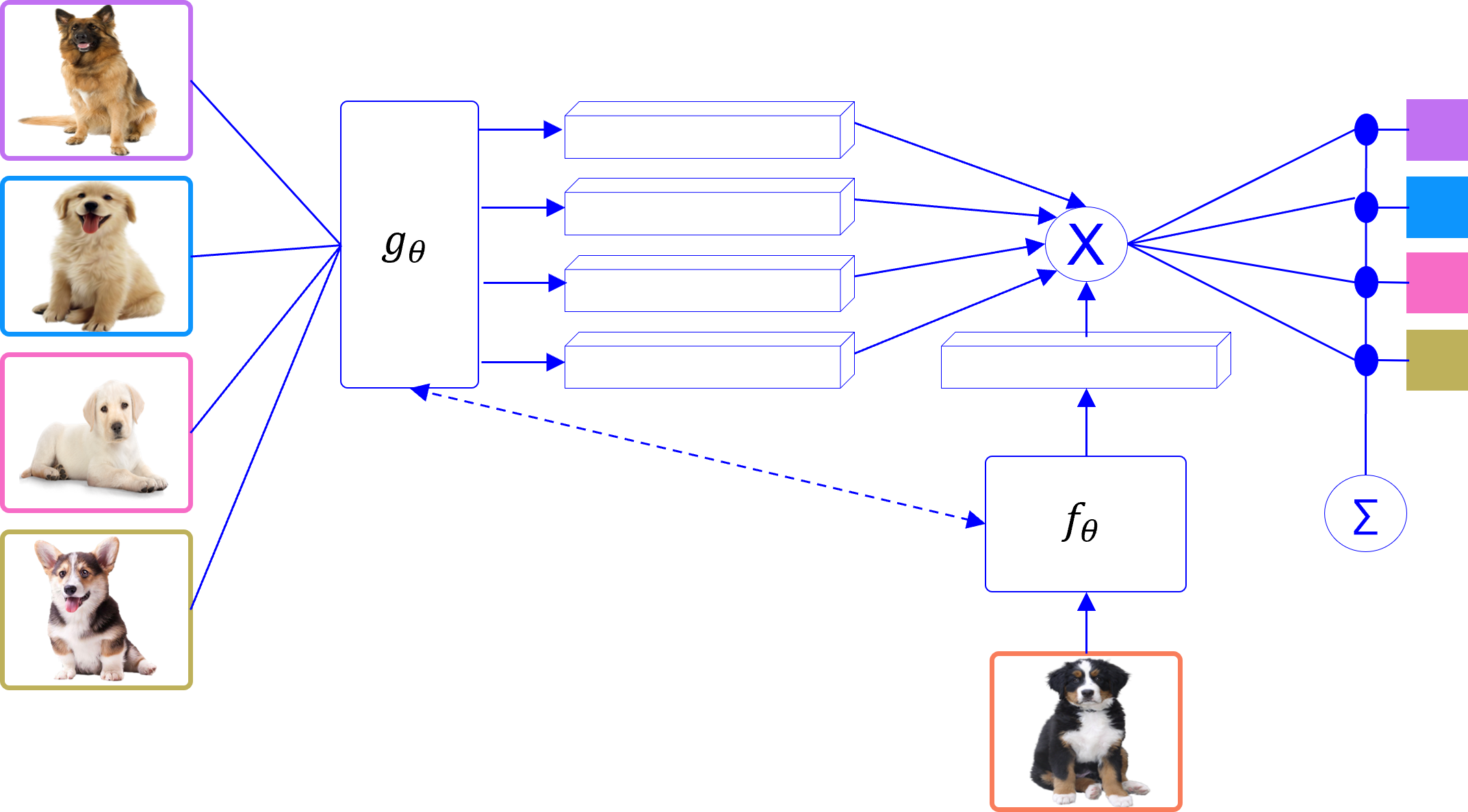}
    \caption{Architecture of Matching Networks.}
    \label{fig_matching_network}
\end{figure}

In Eq.\ref{eq_matching_network_p}, the attention mechanism result is multiplied by the support set labels $y_i$. For this purpose, the support set labels are first converted into one-hot encoded values. The overall flow of the matching network is shown in Figure \ref{fig_matching_network}. As seen in Eq. \ref{eq_matchingNet_attention_mech}, the authors used two different embedding functions, $f$ and $g$, to map data points into embedding for the query $\hat{x}$ and support input $x_i$, respectively. Finally, the proposed network is trained to minimize the prediction error. \cite{chen2021learning} argued that support and query images could differentiate in local features and shapes. They tackled this problem by proposing a cascaded feature matching network (CFMN), which conceptually maps features of the support and query sets to a latent space then reshapes them into a two-dimensional array. Next, it calculates the element-wise similarity of support and query features. Additionally, the support set features maps to a higher space based on element-wise correlation calculated in the previous step. In other words, the final support set features are constructed by taking into account the relations of feature positions, the so-called spatial attention mechanism. This approach allows relevant features to be considered and less important features to be ignored. Finally, CFMN predicts a label for a new sample based on the distance-based similarity score of the support set and a new sample point. \cite{mai2019attentive} proposed another matching network variant called the attentive matching network (AMN), which exhibits two novelties. First, it incorporates a new loss function named complementary cosine loss (CCL) that is a combination of the hardest-category discernment loss (HDL) and cosine-distance metric, aiming to minimize the intra-class difference and maximize the inter-class differences. The second novelty is a new attention mechanism that focuses on the feature level. This module calculates the importance of features based on the standard deviation over classes. In other words, a higher standard deviation over classes means that features are more distinctive between classes. For this purpose, the authors first built class prototypes by class-wise mean of embeddings, then calculated the inter-class standard deviation from each class prototypes. In order for the feature attention module to be differentiable, they proposed a convolutional operation over the attention calculation.

\cite{zhang2019scheduled} argued that a generic matching network randomly selects labels from the base dataset to create training samples, which leads to an over-easy label set. For example, the authors compared a label set {cat, tree, water} to another label set {cat, dog, lion} that is easy and low value in a classification task. To address this issue, they proposed schedule sampling for the matching network, which first samples a random label set, then calculates the difficulty value for the selected label set. The difficulty value is computed based on similarity metrics. The training of matching net started with a low difficulty label set and then continued into a more difficult label set. The authors showed that the performance of the classifier directly relates to the difficulty of the label set.

\subsubsection{Relation Networks}
\begin{figure}[!h]
    \centering
    \includegraphics[width=0.65\linewidth]{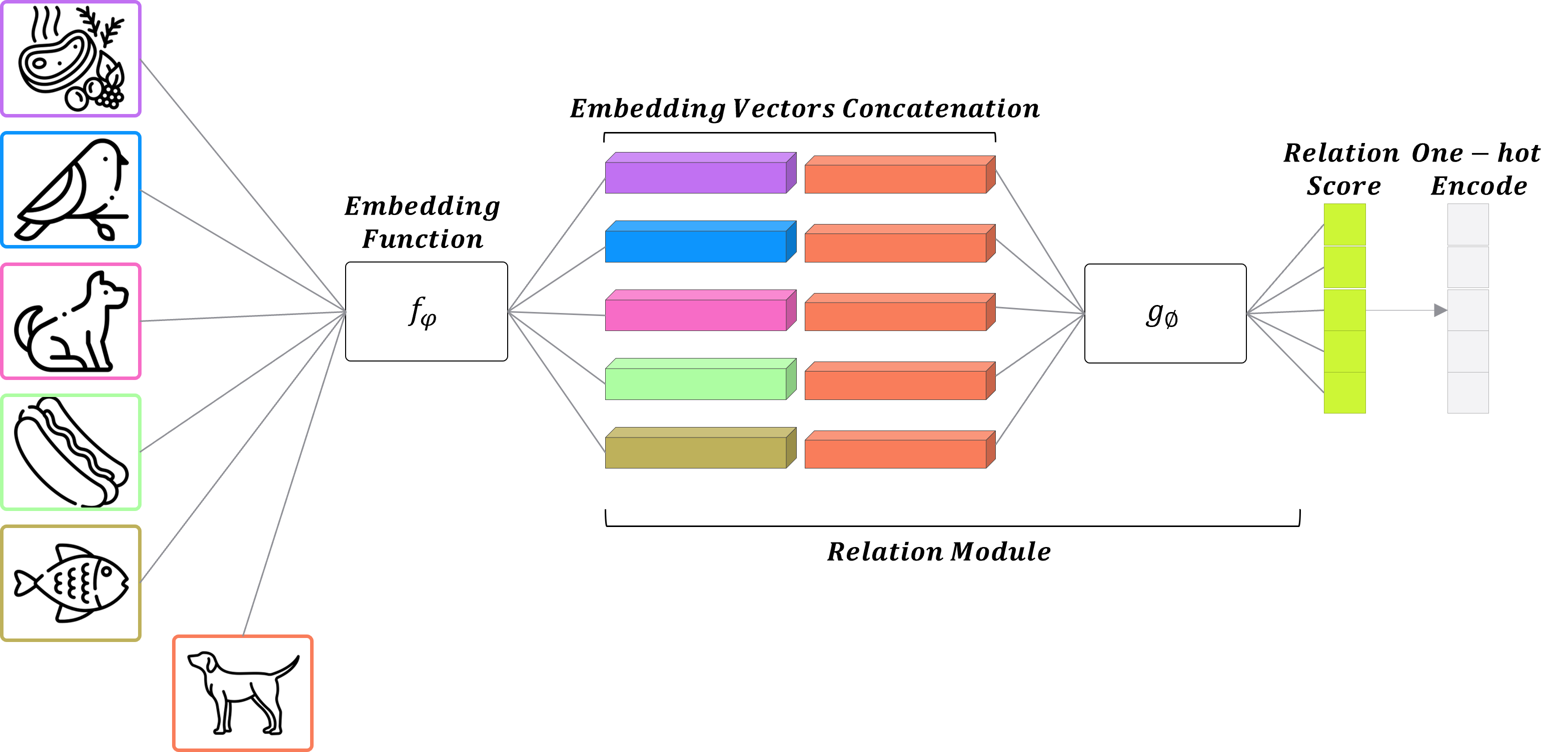}
    \caption{Relation network architecture.}
    \label{fig_relation_network}
\end{figure}

Relation network is another simple yet effective metric-based meta-learner presented by \cite{sung2018learning}. This model has two main components: \romannum{1} embedding function denoted by $f_\varphi$ and \romannum{2} relation function $g_\phi$. According to the flow of the relation network illustrated in Figure \ref{fig_relation_network}, the embedding function $f_\varphi$ maps the support and query sets into the embedding environment. For the k-shot problem where $k>1$, embeddings of the same class in the support set are aggregated through the element-wise summation to construct each class embedding map. The obtained consolidated classes embeddings are then combined with embedding of query samples and fed into the relation function $g_\phi$. The output of the relation function $g_\phi$ is the $N$ (number of classes) similarity score called the relation score that ranges between 0 and 1. While the proposed model is classification task, the authors trained their model using the mean square error (MSE) loss function since they aim to predict a relation score scale between 0 and 1.

With a similar belief to \citep{ji2020improved}, \cite{he2020memory} introduced the memory-augmented relation network (MRN) to investigate interactions of instances. MRN is composed of four components: (\romannum{1}) feature embedding function, (\romannum{2}) information generation component, (\romannum{3}) relation module, and (\romannum{4}) metric-based classifier. The feature embedding function simply converts data points into latent space. The information generation component adds information to samples basically from its k-nearest samples' embeddings. This component is equipped with memory to control embedding propagation. The relation module, similar to a generic relation network, is a convolutional neural network that takes input samples and computes their distances. The last component, the metric-based classifier, predicts a label for an unlabeled query point based on the distance from class centroids, which are defined as the mean of the same class feature representations. This classifier trains to minimize cross-entropy loss.  \cite{liu2020meta} proposed a prototype-relation network by basically combining prototypical networks and relation net. This model has two main components: 1) the prototype component that generates class prototypes, and 2) the relation component, similar to the generic relation net, that computes relations between query points and class prototypes. Another novelty of this work is the introduction of a new loss function that considers inter and intra-class distances. \cite{ding2019multi} developed another variant of relation networks called multi-scale relation networks (MSRN) that has two main components: a feature extractor and metric learner. In convolutional neural networks, low-level layers generate detailed information while the higher-level layers produce more semantic information. The authors argued that combining the feature outputs of different layers may improve representations. The second component, the metric learner, is basically a generic relation network; however, the authors substracted these two components instead of concatenating two multi-scale features. They claimed that subtraction generates a better representation of feature differences. Their model outperforms the generic relation net. \cite{ren2022multi} proposed a multi-local feature relation network (MLFRNet) to improve relation networks. They used random cropping and random dropping to get local features, then mapped to embedding space. Finally, class prototypes are calculated by the mean of each class's local feature embeddings. The authors also introduced a local feature attention module to calculate the relation between local feature prototypes and query samples. Basically, the local feature attention module is responsible for computing the contribution of local features in query sample prediction. Additionally, the authors proposed a new strategy called support-support similarity to dynamically calculate the margin loss of local features. Support-support similarity was designed to compute pair-wise cosine similarity of classes that is used to generate a margin. The authors suggest that adding the margin is necessary for different classes that might have local features with high similarity in order to separate different classes effectively.
\cite{abdelaziz2022multi} argued that the original relation network is not robust to spatial positions of feature maps and may perform poorly in scale variations of compared objects. To address this issue, they extended a new variation of the relation network called the multi‑scale kronecker‑product relation network (MsKPRN), which consists of three main modules. The first module is the encoding module that is responsible for converting different scales of support and query sets into feature space. The output of this module is feature maps with different scales that is organized in a way to feed the same scale support and query feature maps into the next module. The next module is the kronecker‑product (in short KP) module, which calculates the spatial correlation from pair representations by operating the inner product on the same scale support and query feature maps. The obtained spatial correlation is concatenated with the representations to feed into the next module. The third module is the relation network module, which is responsible for learning the relation between the same scale support and query feature maps and producing similarity scores in the range between 0 and 1. Their results demonstrated that the kronecker‑product module improves few-shot classification tasks by incorporating spatial correlation maps into feature maps.

\subsubsection{Other Advanced Methods}
The next advanced metric-based meta-learning algorithm reviewed in this section is the graph neural network (GNN) proposed by \cite{garcia2018few}. The authors argued that their proposed GNN for the few-shot classification problem generalizes Siamese networks, prototypical networks, and matching networks based on their architectures. Conceptually, in their model, each task $T_j$ is approached as a fully-connected graph $G=(V,E)$, where $V$ and $E$ are denoted as nodes and edges, respectively. Nodes are associated with support set embeddings concatenated with one hot encoded label. Edges correspond to tasks' similarities, which are trainable. For a new query point, i.e. an unlabeled data point, a uniform distribution over all labels is used as label concatenation to query embedding. Finally, GNN was trained to minimize the cross-entropy loss. Results show that GNN outperforms Siamese networks, prototypical networks, and matching networks on several benchmark problems. \cite{li2019few} proposed the global class representation (GCR) model, which achieved the best performance on benchmarks and outperformed many well-known meta-learning algorithms. The idea of GCR is to engage novel classes with base classes at the first stages of training, thus the reason for naming the model global class. Concretely, global class representation is first created from the whole class set, called $C_{Total}$, which is composed of both base and novel classes. Next, in each episode, $N$ random classes and $n_s$ random samples per each class are selected to form the support set. Then, episodic class representation is built based on the support set. Here, class representation is analogous to class prototypes in prototypical networks. As one novelty of this study, the authors proposed a new component called the registration module. Basically, the registration module is responsible for calculating the similarity between each episodic class representation and global class representations, or the so-called registration loss. GCR minimizes both registration loss and classification cross-entropy loss to achieve optimal global classes. GCR predicts a label of a new sample by comparing it to global class representations. \cite{han2020multi} disputed that most metric-based meta-learners use standard similarity measures, such as cosine similarity and Euclidean distance. They support that those metrics could not capture non-linear relations between samples. To address this issue, they introduce a multi-scale feature network (MSFN), which has three main components. The first component is the feature extractor that extracts multi-scale features from the support set. The next component is the label feature that is responsible for learning classes representations (analogous to class prototypes). The label feature basically is a network that takes a concatenation of support features from the previous component as the input and generates class representations as the output. The last component is the metric-component, which principally is the relation network that computes the matching degree between a new query feature map and class representations. Through numerical analysis, the authors demonstrated that multi-scale features and non-metric similarity measures improve the overall model performance. 

\begin{figure}[!h]
    \centering
    \includegraphics[width=0.5\linewidth]{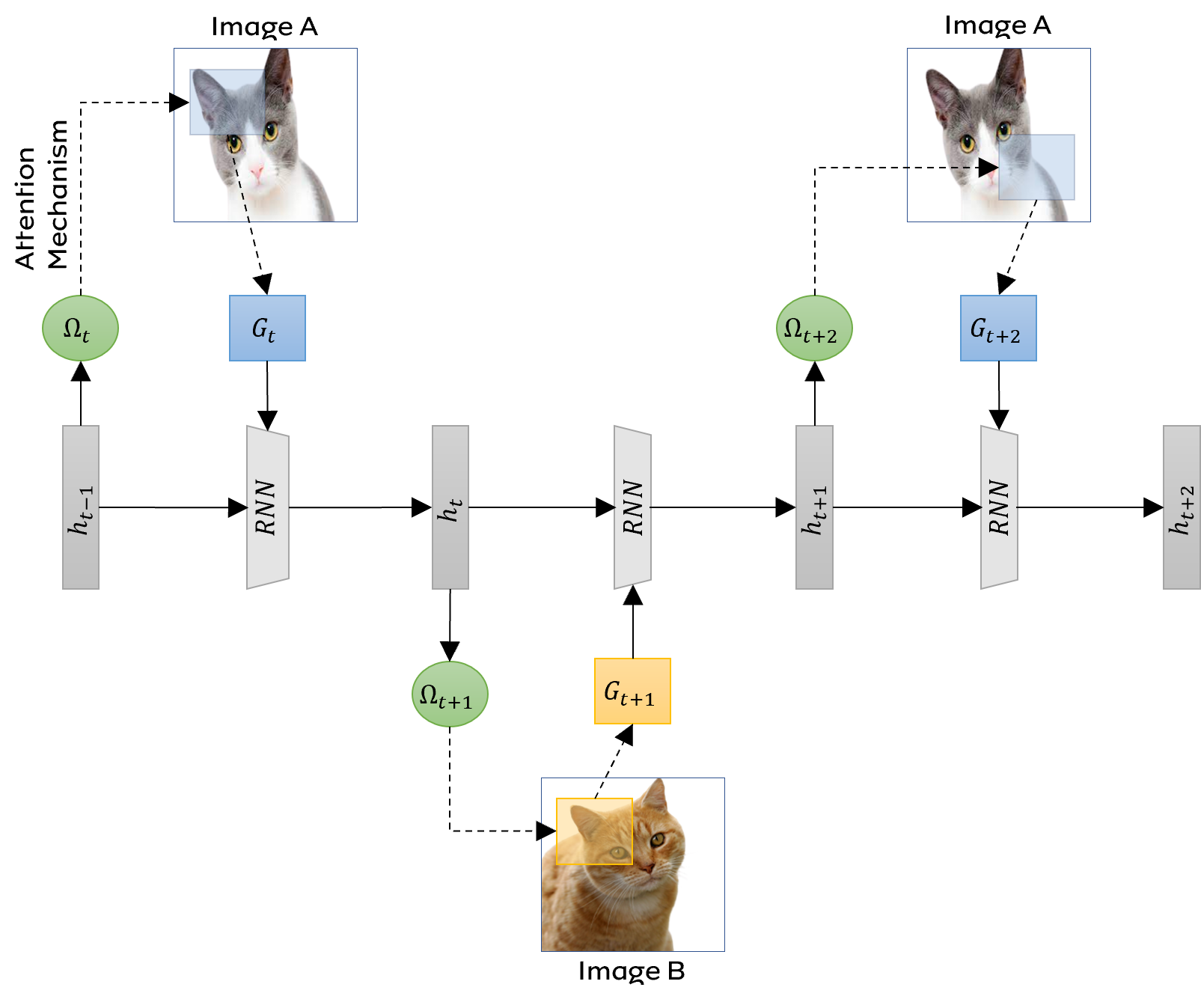}
    \caption{Attentive recurrent comparators (ARCs) scheme.}
    \label{fig_ARCs}
\end{figure}

\begin{figure}[!h]
    \centering
    \includegraphics[width=0.5\linewidth]{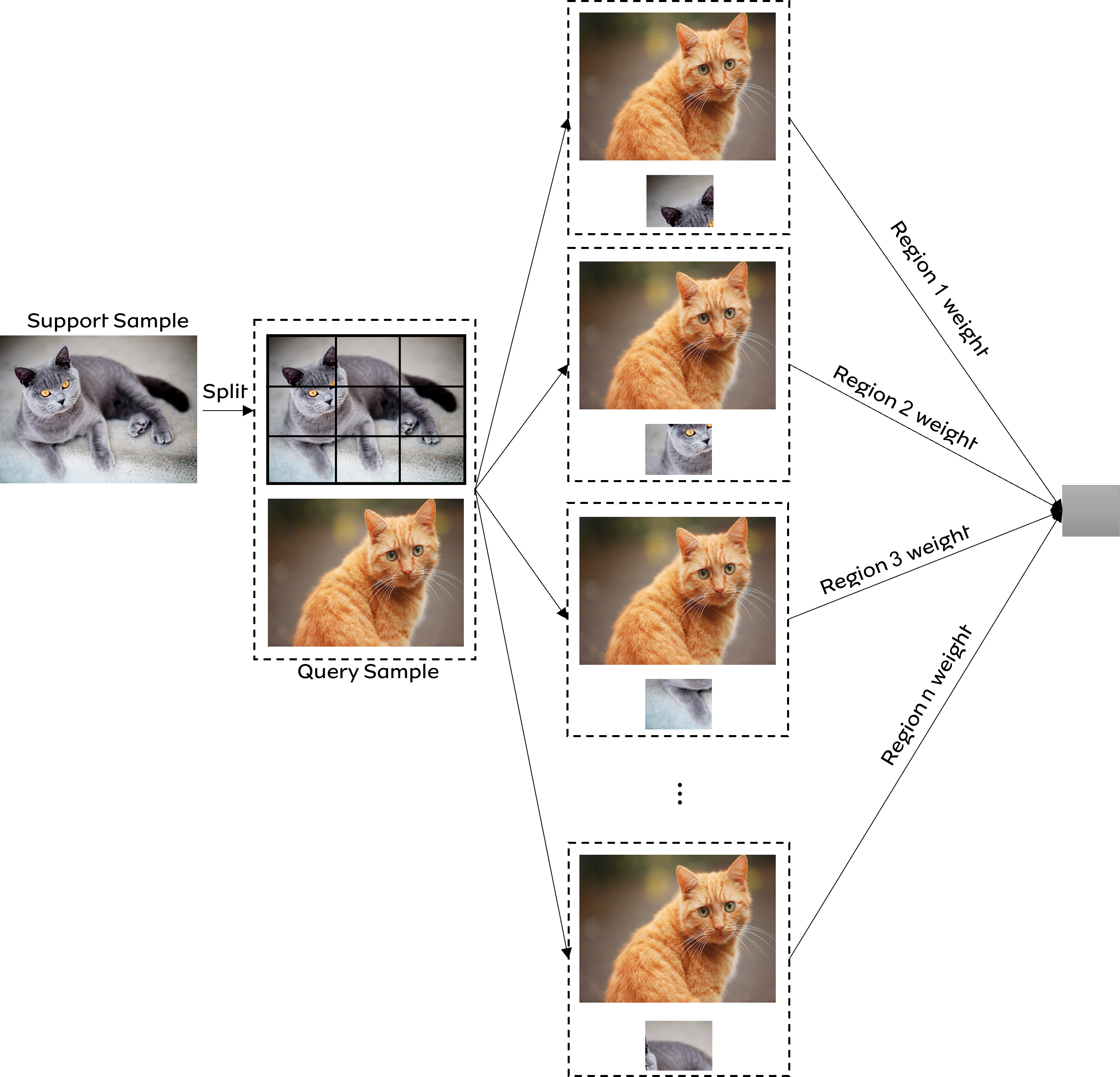}
    \caption{Comparison support sample's parts with query sample in Region comparison network (RCN).}
    \label{fig_RCN}
\end{figure}

\cite{shyam2017attentive} argued that humans differentiate two objectives by recurrent multiple observations, and during each observation, a specific feature is inferred. This human-wise approach is quite different from Siamese networks. \cite{shyam2017attentive} proposed attentive recurrent comparators (ARCs), mimicking humans. ARCs consists of two components: (\romannum{1}) recurrent neural network (RNN) controller, and (\romannum{2}) attention mechanism. The attention mechanism is designed in a way that concentrates on a specific region of the image. Figure \ref{fig_ARCs} demonstrates the scheme of ARCs, in whichthe attention mechanism gets a glimpse $G_t$ from the image $I_t$ at each step (Eq.\ref{eq_ARCs_glimpse}):

\begin{equation}
    G_t \leftarrow (I_t, \Omega_t)
    \label{eq_ARCs_glimpse}
\end{equation}
where $\Omega_t$ is the attention glimpse parameter, denoting the location and size of the glimpse. Basically, $\Omega_t$ is computed based on the previous hidden state ($h_{t-1}$) of the RNN controller (Eq.\ref{eq_ARCs_omega}):

\begin{equation}
    \Omega_t = w_t h_{t-1}
    \label{eq_ARCs_omega}
\end{equation}
where $w_t$ is the projection matrix necessary to map $h_t$ to the number of attention parameters. Next, the hidden state is obtained via Eq.\ref{eq_ARCs_hiddenstate}:

\begin{equation}
    h_{t} \leftarrow  RNN(G_t, h_{t-1})
    \label{eq_ARCs_hiddenstate}
\end{equation}
where $RNN(\cdot )$ is the RNN controller update function. In short, ARCs recurrently observe different partitions of two images over a finite number of representations, so the model can focus better on the important context of each image. While the presented model outperforms Siamese networks, it can be quite computationally expensive due to sequential operations. Similarly, \cite{xue2020region} believed that humans differentiate objects based on partitioning images and comparing them with the prototypical part of each image category. Accordingly, they proposed a new metric-based method called the region comparison network (RCN). The aim of RCN, as demonstrated in Figure\ref{fig_RCN}, is to compare the whole query image with different partitions of support samples. RCN consists of three main components: (\romannum{1}) feature extractor, which is responsible for producing image representations; (\romannum{2}) region matching, which is designed in a way to calculate the similarity scores between support and query samples; and (\romannum{3}) explain network, which is used to obtain the final classification. To be more specific, the output of the first step is decomposed into several region vectors. Then, cosine similarity between support set region vectors and query set representation is computed. Subsequently, global max-pooling is applied to pick the most important information in the region similarity vector. The explain network takes region similarity maps from the previous step and region weight to make a classification. In each task, the important region may change. For example, in classifying dogs and birds, the important regions are the dog's head and bird's wings. The explain network uses task-specific region weights produced by the region meta-learner, which is designed to take a concatenation of support set representations and query set representations and produce task-specific region weights. The most salient aspects of RCN are its interpretability and quantitative explainability of classification tasks in the few-shot setting. 

\cite{hilliard2018few} aimed to explore the effects of pre-defined similarity metrics, such as Euclidean and cosine distances, on existing metric-based methods and proposed a new method called metric-agnostic conditional embeddings (MACO). The architecture of MACO is demonstrated in Figure\ref{fig_MACO}. Basically, MACO is composed of four stages: (\romannum{1}) feature stage, (\romannum{2}) relation stage, (\romannum{3}) conditioning stage, and (\romannum{4}) classifier stage. The first stage is responsible for obtaining a feature vector from input images, which is simply a CNN-based network with a fixed structure for both the support and query sets. The next stage is designed to produce a single vector for each class. This class-specific single vector is analogous to the class prototype. The authors used the relational network (proposed by \cite{santoro2017simple}) to generate the single vector class representations. The relational network is comprised of fully connected layers (instead of convolutional layers) followed by batch normalization and ELU activation function using pair-wise image comparisons within each class. The single vector class representation would be the average output across all comparisons (Eq.\ref{eq_MACO_single_vector_class}):

\begin{figure}[!h]
    \centering
    \includegraphics[width=0.4\linewidth]{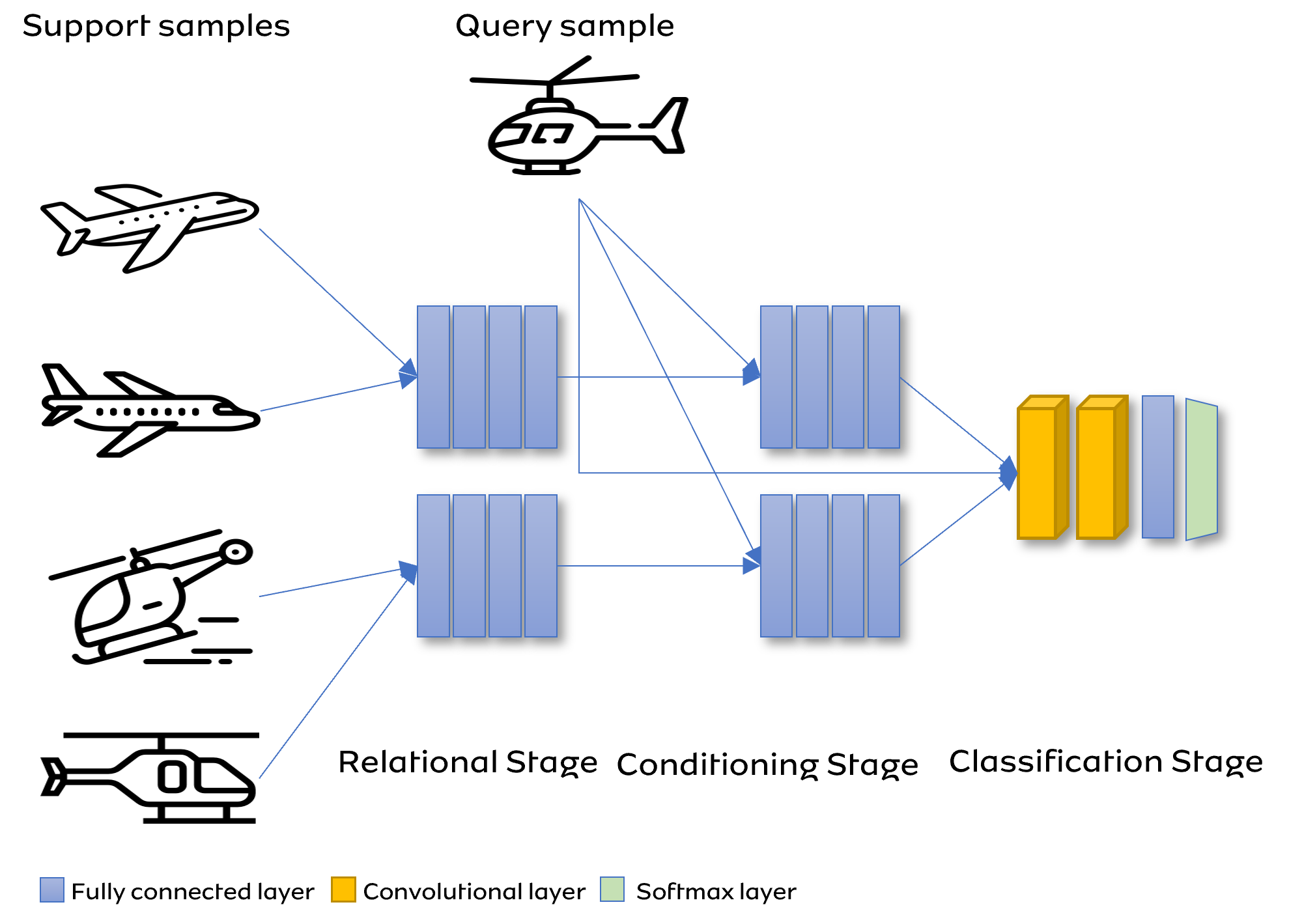}
    \caption{Metric-agnostic conditional embeddings (MACO) scheme.}
    \label{fig_MACO}
\end{figure}

\begin{equation}
    R_\rho (s_k) = \frac{1}{\binom{n}{2}} \sum_{(x_i, x_j)\in s_k}^{} g_\rho(x_i, x_j), \ i\neq j
    \label{eq_MACO_single_vector_class}
\end{equation}
where $R_\rho$ is the single vector of support set class $k$; $n$ is the number of images in class $k$; $g_\rho$ is the relational network parameterized with $\rho$; and $x_i, x_j$ are images representations within support class $k$. The conditioning stage uses a network similar to a relational network that takes the concatenation of class vectors and query images. The output of this stage is the modified class representation in the context of the query image. The last stage is a neural network that takes the modified class vectors and query images as input and produces Softmax classification. This network is composed of convolutional layers followed by fully connected layers. In short, MACO is a novel model in which a learnable classifier supersedes the pre-defined metrics and modified class vectors (or prototypes) conditioned to the query image context. Similarly, \cite{xue2020relative} developed a learnable network instead of pre-defined similarity metrics termed relative position and map network (RPMN). Basically, RPMN consists of three components: (\romannum{1}) feature extractor, (\romannum{2}) relation position network, and (\romannum{3}) relative map network. The relative position network aims to produce different weights to the position of feature embeddings, and the relative map network is analogous to the original relation network but has been modified in such a way to compare two single maps from support and query feature representations. The relative map network is trained to learn task-specific embedding models. Moreover, the relative map network is equipped with a fully connected head, which is responsible for computing the final similarity score between the support and query sets. An important aspect of RPMN is the introduction of the relation position network as a new attention mechanism, allowing for more comparisons of effective feature maps, and the relative map network as a learnable similarity metric network.

\subsection{Memory-based Methods}
\begin{figure}[!h]
    \centering
    \includegraphics[width=0.4\linewidth]{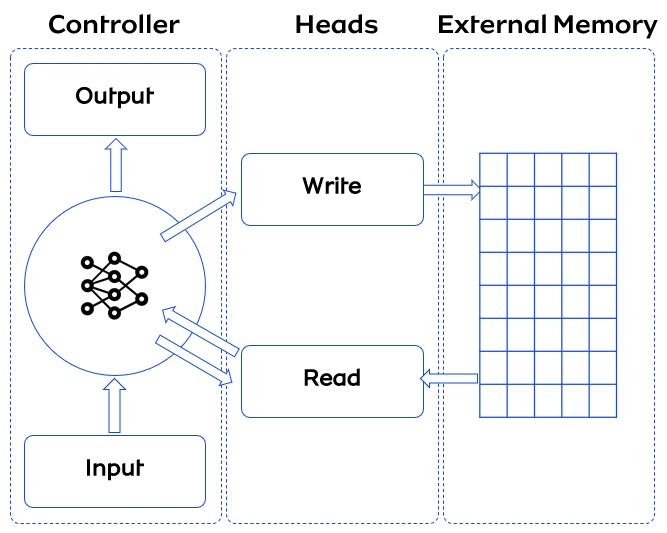}
    \caption{Neural Turing Machine scheme.}
    \label{fig_NTM}
\end{figure}

Memory-based meta-learning techniques have a more dynamic mechanism since they are adaptive to the presented tasks. Principally, memory-based meta-learners own the memory component (internally or externally) that enables them to retrieve information from previous inputs. The memory-augmented neural networks (MANN) is a well-known and effective memory-based meta-learner proposed by \cite{santoro2016meta}, which aims to use external memory to improve task adaptation utilizing the Neural Turing Machine (NTM) \cite{graves2014neural}. On a high level view, NTM is a neural network equipped with external memory to retrieve information. More specifically, NTM contains two main components: (\romannum{1}) controller and (\romannum{2}) memory bank. Figure \ref{fig_NTM} shows the interaction of the controller and memory bank. Similar to a standard neural network, NTM receives inputs and predict outputs. In contrast to a standard neural network, NTM communicates with the memory bank through selective read and write operations, which the authors named as \textit{head} analogous to the Turing machine. In order to NTM be trainable, including the way memory bank is utilized, the authors developed blurry read and write operations that should be differentiable to use backpropagation. For this purpose, read and write operations interact with distribution over memory locations instead of a single location. Specifically,  the read operation simply reads values from the memory bank, where the value is basically the neural network parameters (i.e., weights). The memory bank is simply a two-dimension matrix. The write operation is more complex than read and includes two modules called erase and add, which erases information that is no longer required from a memory cell and adds a new operation to the memory, respectively. 

To produce the parameters of the NTM neural network, the authors proposed two different addressing mechanisms. First, the content-based approach uses the similarity of the current value and controller's value. While the content-based is quite straightforward, the location-based approach is employed to iterate over different locations rather than content, which is essential for generalization. \par

\begin{figure}[!h]
    \centering
    \includegraphics[width=0.3\linewidth]{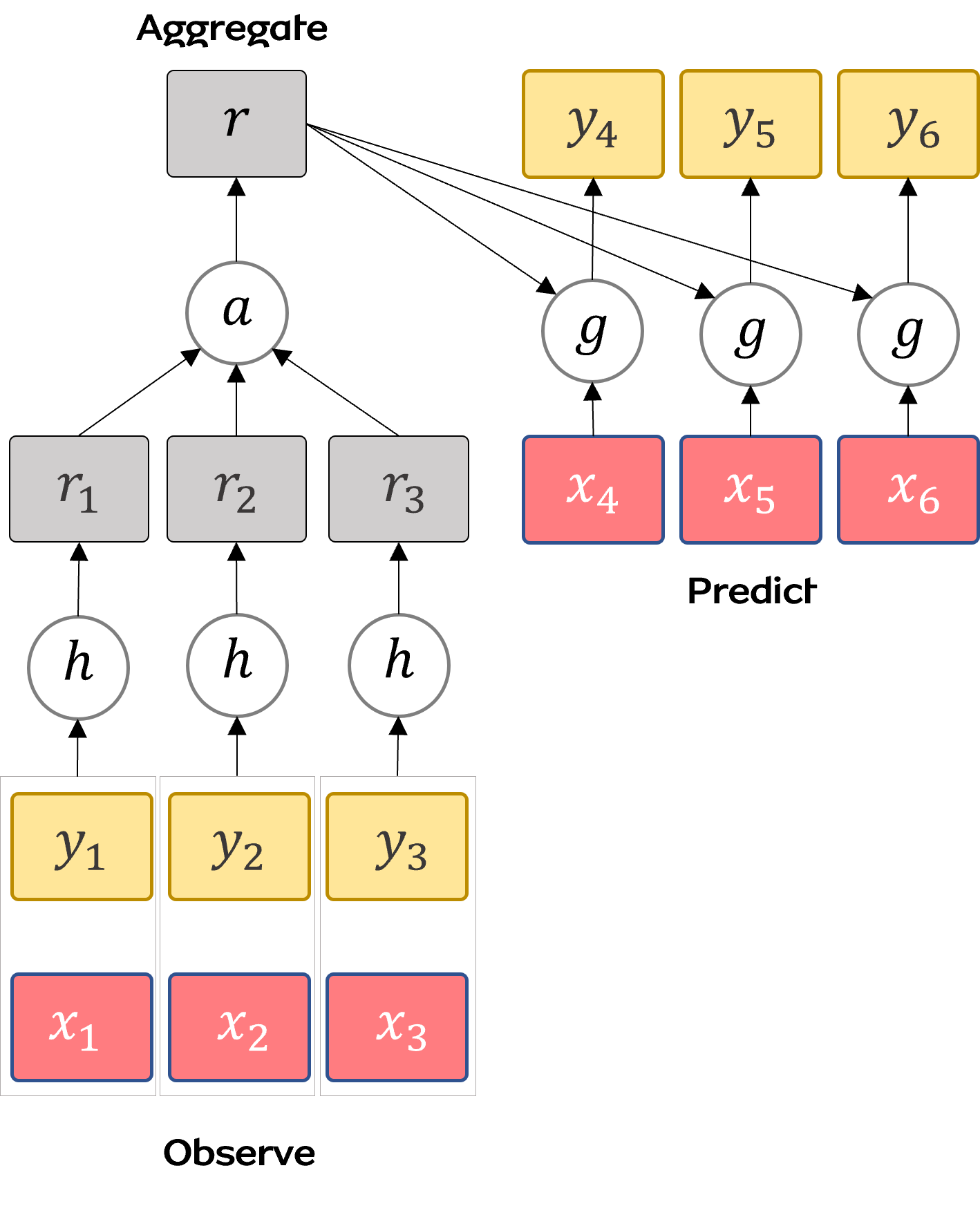}
    \caption{Conditional Neural Processes scheme.}
    \label{fig_CNPs}
\end{figure}

\noindent \cite{santoro2016meta} proposed a new addressing mechanism called least recently used access (LRUA). In MANN, the read operation is performed through the content-based addressing mechanism, and the write operation uses LRUA to write the most recently used memory location. MANN accumulates information of a coupled representation-class label in external memory through the explained mechanism, which enables it to retrieve the information later for classification [specifically a sample from seen class is given - please reword this last part for clarity].  
\cite{mishra2017simple} proposed a new memory-based meta-learner model called simple neural attentive meta-learner (SNAIL) that combines temporal convolution and soft attention mechanism that acts as a memory compared to the external memory of MANN. The temporal convolution serves as high band-width memory access, and the soft attention allows access to a specific experience. The combination of these two components leads to better leveraging of information from past experiences. \cite{garnelo2018conditional} proposed conditional neural processes (CNPs) that, in a similar fashion to SNAIL, do not employ external memory. On a high level, CNPs consists of a meta-learner and task-learner. Figure \ref{fig_CNPs} shows the general architecture of CNPs. In this figure, $h$ is an embedding function that generates representations $r_i$ from the input and ground truth (labels) from the support set $S$. Operator $a$ is the aggregation function that produces a single representation from $r_i$, and $g$ is the neural network (task learner) that predicts labels for a new sample from the query set based on the output of $a$. As seen in Figure \ref{fig_CNPs}, the meta-learner produces a memory value by aggregating representations of the support set through operator $a$. Then, the label of new inputs is predicted by processing the aggregated representations. In other words, CNPs use concise representations of seen classes to make predictions.

\cite{tran2019memory} proposed a memory augmented matching network by combining MANN and matching network. The authors argued that if support data distribution and skewed, constructing a prototype by mean of embeddings leads to biased class prototypes. To support this argument, they developed weighted class prototypes by incorporating the distances of class-wise samples. The weight of sample $i$ in each class is calculated as the inverse of the total distance of sample $i$ to all other samples in that class. The results showed that weighted class prototypes lead to remarkable improvement.

\subsection{Learning-based Methods}
This work categorizes the three groups of \textit{learning the initialization}, \textit{learning the parameters}, and \textit{learning the optimizer} as a subset of a larger group \textit{learning-based methods}. In previous surveys, three groups were introduced as model-based methods or optimization-based methods, but these methods are mechanically different. This work refers to these three groups as learning-based methods because a network (meta-learner) is trained to learn shared global initializer (learning the initialization), to learn the learning rate (learning the optimizer), or to learn the network's parameters for the base learner (i.e., the model that predicts for unseen tasks). The following subsections discuss each group in detail. 

\subsubsection{\textbf{Learning the initialization}} \hfill\\

\begin{figure}[!h]
    \centering
    \includegraphics[width=0.28\linewidth]{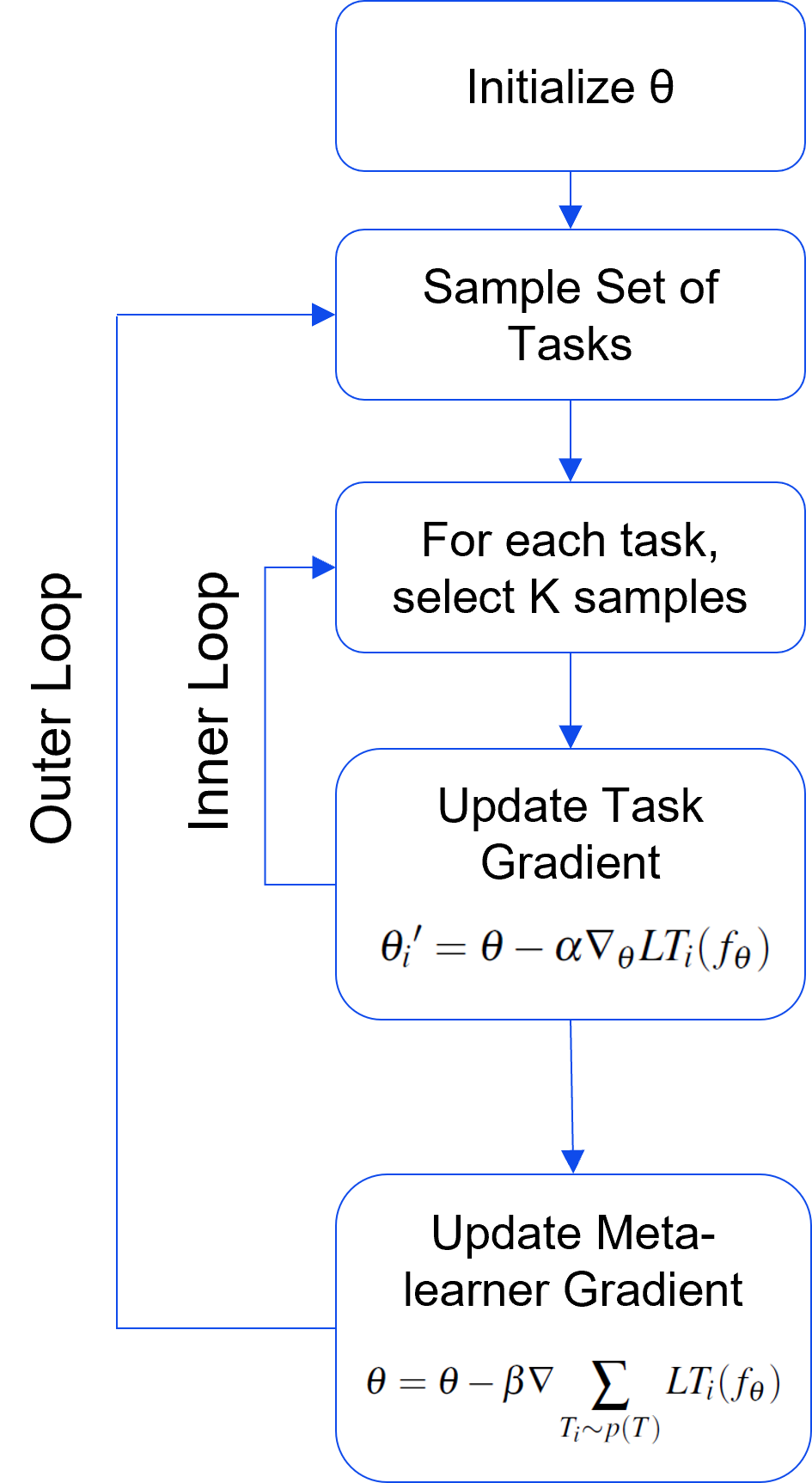}
    \caption{Model agnostic meta-learning workflow.}
    \label{fig_MAML_Workflow}
\end{figure}

In a standard neural network, weights are initialized randomly, then after calculating loss, optimal weights are obtained by minimizing loss through the gradient descent process. The learning an initialization algorithms seeks to generate a better shared global initializer learned from the support set so that a new unseen task can learn quickly with fewer data points and less optimization steps. The model-agnostic meta-learning (MAML) is a pioneering learning the initialization algorithm proposed by \cite{finn2017model}. Intuitively, MAML aims to learn a model's parameters in such a way that a gradient-based learning rule on a new unseen task $T$ progresses quickly. Figure \ref{fig_Learning_an_initialization_MAML} shows how MAML can lead to improvement on the direction of the gradient. In this figure, three different related tasks are represented with optimal parameters ${\theta}'_1$, ${\theta}'_2$ and ${\theta}'_3$, where $\theta$ is the randomly initialized parameter. By shifting $\theta$ to a new position closer to ${\theta}'_1$, ${\theta}'_2$ and ${\theta}'_3$, these optimal parameters can be achieved quickly with fewer steps. The general architecture of MAML consists of a model denoted by function $f_\theta$ with parameter $\theta$. The model parameter $\theta$ is updated by Eq. \ref{eq_MAML_1} :

\begin{figure}[!h]
    \centering
    \includegraphics[width=0.6\linewidth]{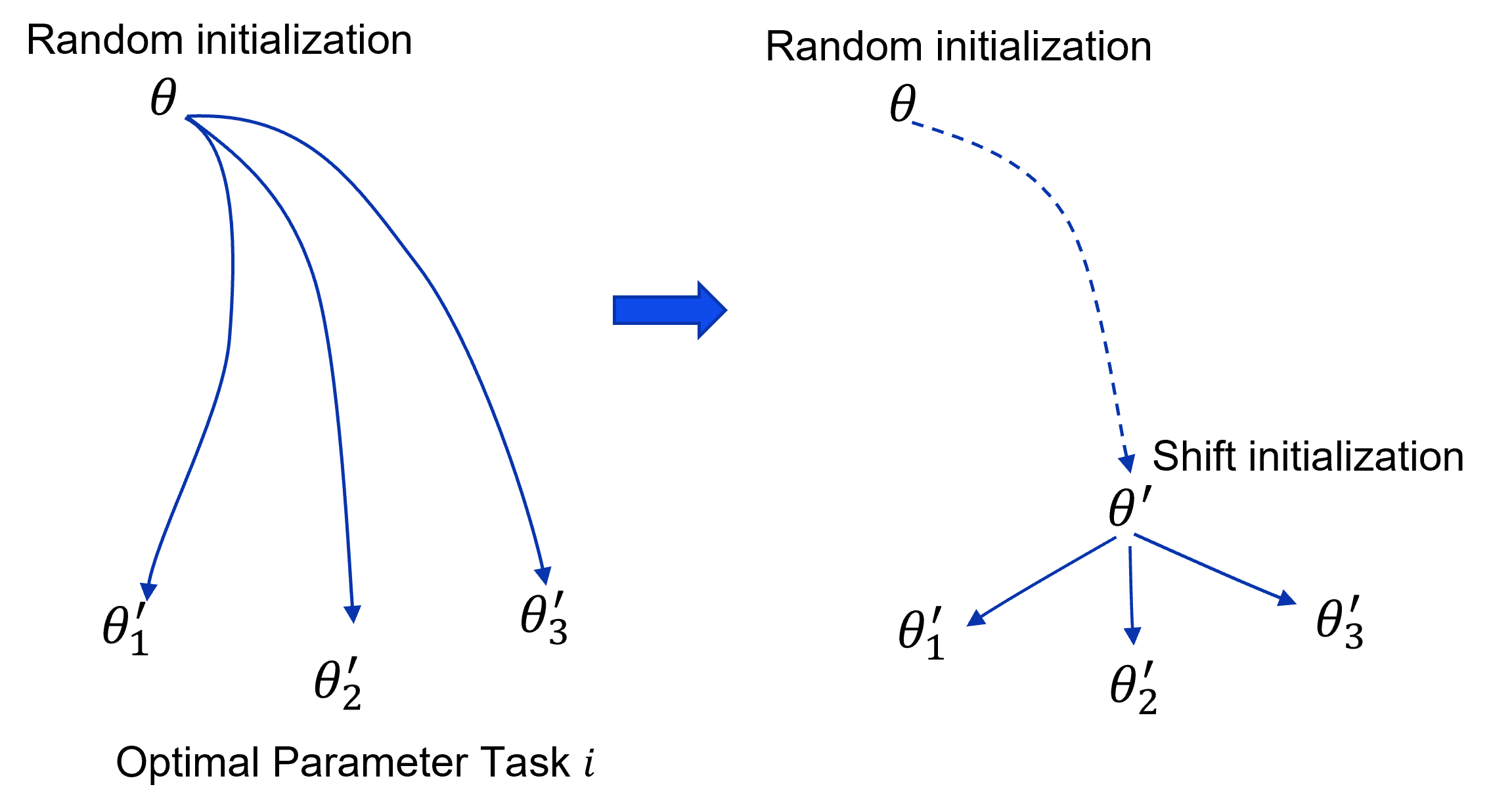}
    \caption{Schema learning initialization, Model agnostic meta-learning for fast adaptation.}
    \label{fig_Learning_an_initialization_MAML}
\end{figure}

\begin{equation}
    {\theta_i}' = \theta  - \alpha \nabla_\theta\mathfrak{L}T_i(f_\theta)
    \label{eq_MAML_1}
\end{equation}
where $\mathfrak{L}T_i(f_\theta)$ is a gradient of loss for task $T_i$; and $\alpha$ is a hyper-parameter. In this step (inner loop), the optimal parameter $ {\theta_i}'$ for each task $T_i$ is obtained. In the next step (outer loop), the randomly initialized parameter $\theta$ is updated on a new set of tasks by Eq.\ref{eq_maml_teta_update}. The flow of inner and outer loops update rule is shown in Figure \ref{fig_MAML_Workflow}. Thus, parameter $\theta$, initialized randomly, shifts to an optimal position where fewer gradient steps are required for the next batch of tasks. 
\begin{equation}
    \theta = \theta  - \beta \nabla \sum_{T_i\sim p(T)} \mathfrak{L}T_i(f_\theta)
    \label{eq_maml_teta_update}
\end{equation}
where $\beta$ is a hyper-parameter; $p(T)$ is distribution over tasks $T_i$ and $\mathfrak{L}T_i(f_\theta)$ is the gradient for each new task $T_i$.

\begin{figure}[!h]
    \centering
    \includegraphics[width=0.6\linewidth]{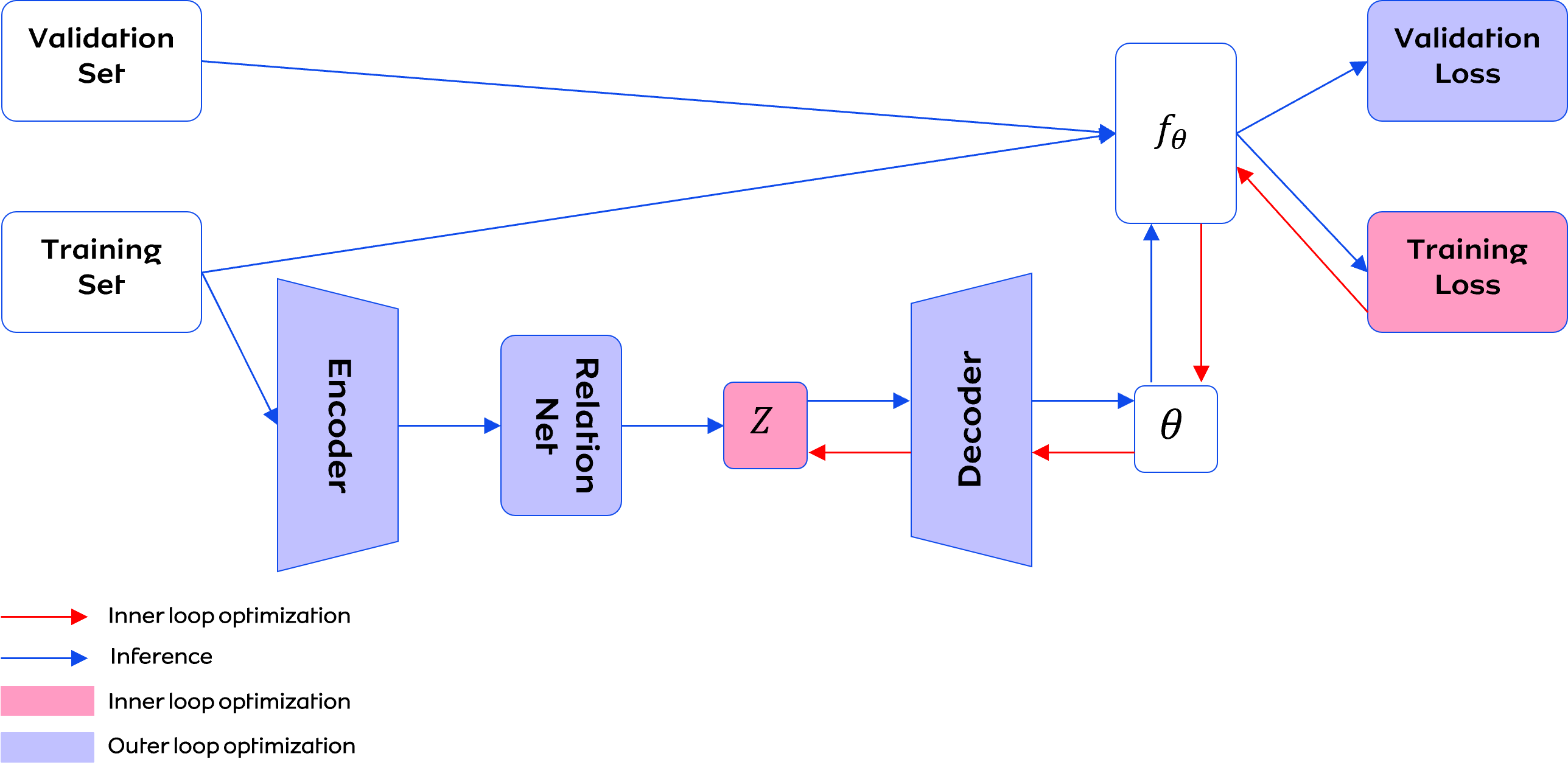}
    \caption{LEO scheme.}
    \label{fig_LEO_Architecture}
\end{figure}

Continuing their previous work,\cite{finn2018probabilistic} proposed probalistici MAML, inspired by \cite{grant2018recasting} who interpreted MAML as an approximate inference for the posterior distribution of parameters of the support set. The proposed model, termed probabilistic latent model for incorporating priors and uncertainty (PLATIPUS), was able to capture uncertainty via variational inference. Numerical experiments indicated that PLATIPUS can effectively be used for a diverse family of tasks with multi-modal task distributions. In another tweak to MAML, \cite{yoon2018bayesian} proposed Bayesian Model-Agnostic Meta-Learning (BMAML) to learn uncertainty, which principally combines the stein variational gradient descent (SVGD) with the MAML algorithm. Their experiments demonstrated that this model is robust to overfitting during the meta-update phase. 
To alleviate the difficulties of learning in high-dimensional parameter space such as such as those faced by MAML, \cite{rusu2018meta} proposed a novel algorithm called latent embedding optimization (LEO). Similar to MAML, LEO consists of an inner loop where the task-specific values are obtained and an outer loop where global shared initializations are updated. Theoretically, LEO improves generalization performance by lowering the dimension of embedding space by using a combination of an encoder and relation network. The workflow of LEO is shown by Figure \ref{fig_LEO_Architecture}. The encoder produces hidden codes from the support set. Then, hidden codes are concatenated pairwise and fed into a relation network, resulting in a probability distribution over latent codes in a lower dimension. Finally, the decoder generates task-specific initial parameters that are differentiable to backpropagate for adaptation.

While meta-learning algorithms have a remarkable performance on benchmark few-shot problems, they are not robust to the adversarial samples \citep{goldblum2020adversarially}. \cite{zintgraf2018caml} proposed a modified MAML model called context adaptation for meta-learning (CAML) by introducing context parameters. Context parameters essentially are task-specific parameters updated in an inner loop. In each iteration of the adaptation process, context parameters are initialized with zero value. This modification leads to better memory management and faster adaptation. Additionally, through the numerical examples, the authors demonstarted that CAML outperforms MAML in regression tasks. \cite{yin2018adversarial} proposed a network robust to adversarial samples by training a MAML model on both clean and adversarial samples. In the developed ADversarial Meta-Learner (ADML) model, inner and outer gradient updating is performed regarding correlations between clean and adversarial examples. This research opens up a new direction in meta-learning. However, the potential problem of meta-learning models, such as MAML, with learning the initialization is that they only learn initializers, which can be biased toward some tasks and leads to poor generalization, specifically on unseen tasks. Addressing this issue, \cite{jamal2019task} proposed a task-agnostic meta-learner (TAML) model, in which two approaches are used to train a model unbiased over tasks. The first approach, "entropy maximization reduction," makes a random guess over predicted labels, so the initial model will have a large entropy. Then, entropy is combined with the meta-learner loss to minimize the entropy, or in other words make the model more confident about the predicted labels. The limitation of the entropy-based approach is that it can only apply to discrete labels. Thus, the authors proposed the second approach, "inequality-minimization," inspired by economic inequalities theory. They used a large group of statistics, including the Gini-coefficient, Theil-index, and variance of logarithm, to compute the tasks bias. They presumed the loss of the initial model of each task $T_i$ as income inequality, which was minimized by the meta-learner. 

\begin{figure}[!h]
    \centering
    \includegraphics[width=0.45\linewidth]{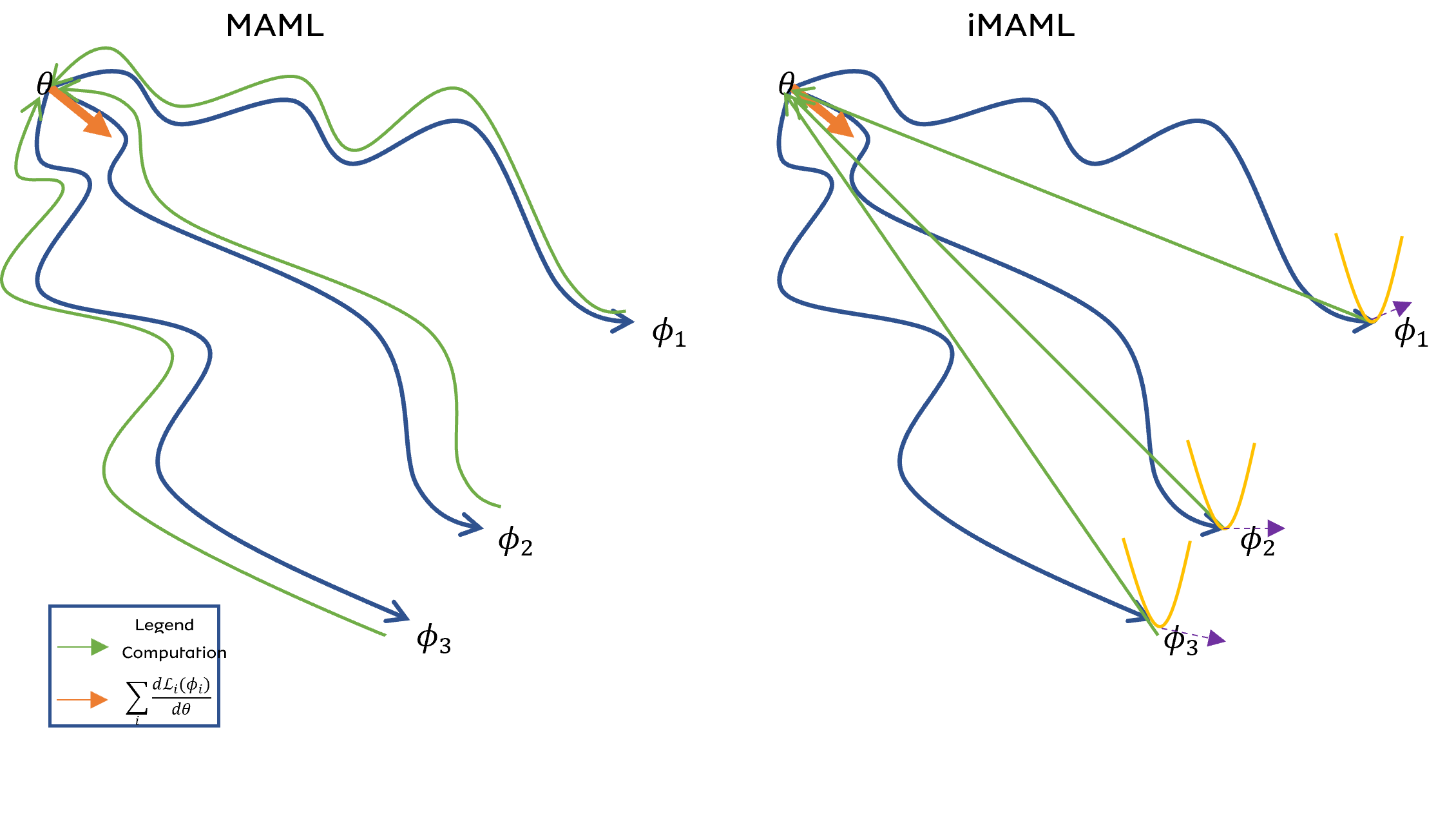}
    \caption{Implicit MAML (iMAML) approach. While MAML differentiates through the optimization path, iMAML estimates local curvature for the exact meta-gradient}
    \label{fig_imaml}
\end{figure}

As can be seen in Eq. \ref{eq_MAML_1} and Eq. \ref{eq_maml_teta_update}, MAML has two hyperparameters, $\alpha$ and $\beta$, that are required to be tuned, which can be expensive. \cite{behl2019alpha} alleviate the tuning burden of MAML hyperparameters by utilizing the hypergradient descent (HD) algorithm (proposed by \cite{baydin2017online}). Theoretically, HD adapts hyperparameters by backpropagating on the hyperparameters (here, $\alpha$ and $\beta$) in conjunction with the original gradient descent iterations. \cite{oh2020boil} believed that representation change plays an important role in the performance of MAML. Theoretically, the body of the embedding function in the MAML inner loop is almost fixed and the embedding function's head is responsible for updating the features. This mechanic led to only small changes in representations during MAML inner loop task updates, which authors termed as representation reuse. The authors proposed a new variant of the agnostic meta-learner model called body-only update in inner loop (BOIL). Contrary to the previous belief of MAML, in their model, the body of the embedding function updates features, leading to remarkable changes in representations, phrased representation change. Empirical results on benchmark datasets showed that the learning body of the embedding function in the inner loop improves MAML performance. \cite{baik2020learning} investigated the possible disagreement of the initialization location between tasks as well as layers of neural networks. They argued that sharing an initialization between conflicting tasks, i.e, tasks with unaligned gradient directions, degenerates the optimization landscape. To solve this issue, they proposed a new technique called learn to forget (L2F), which is designed to selectively forget prior knowledge for each task. For this purpose, they proposed a learnable parameter for each layer that is computed through gradients of the support set batch of tasks. Through numerical analysis, the authors showed that the proposed attenuation parameters lead to a stable performance improvement of MAML. \cite{grant2018recasting} empirically showed that MAML can be cast within a hierarchical Bayesian model to estimate task-specific parameters and, thus, proposed another variation of MAML called Laplace approximation for meta-adaptation (LLAMA). More specifically, LLAMA is an extension of MAML that incorporates uncertainty about task-specific parameters, so instead of point estimation, Laplace approximation is applied to obtain the probability distribution over parameter tasks $T_i$. While LLAMA resulted in a more accurate estimate than MAML's point estimation, Laplace approximation might be quite inaccurate. \cite{rajeswaran2019meta} proposed another extension to MAML called implicit MAML (iMAML). Figure \ref{fig_imaml} illustrates the difference between MAML and iMAML. While  MAML differentiates the meta-gradient through the optimization path, the proposed iMAML estimates local curvature for the exact meta-gradient and is also more memory-efficient since there is no need to differentiate throughout the optimization path.

\subsubsection{\textbf{Learning the parameters}} \hfill\\
In this group of algorithms, the parameters of task-specific networks (i.e., base learner) are generated by another learner (i.e., meta-learner) instead of being initialized randomly, leading to a faster adaptation for new tasks. One early work in this category was conducted by \cite{edwards2016towards}, who proposed neural statistician, a novel extension of the variational auto-encoder that summarizes dataset and learn dataset statistics. In simple words, a neural statistician generates summary statistics, including mean and variance, to provide a generative model that can be trained unsupervised and is used to estimate the posterior distribution of task-specific parameters. However, the major limitation of the neural statistician is that its performance depends on a large dataset. 
\cite{munkhdalai2017meta} proposed a meta-learner model called MetaNet to quickly parameterize a predictive model (i.e., a neural network) by processing meta information. MetaNet consists of two main modules: the base learner operates at the level of tasks and provides the meta-learner with information including loss gradients; and the meta-learner principally generates weights quickly in a task-agnostic setting by processing received information. These generated weights are employed to leverage both the meta- and base learners. In other words, the meta-learner parameterizes itself and the base learner. Here, the meta-learner benefits an external memory to store the support set inputs, their representations, and generated example level weights. Prediction over the query set is made based on comparing the query sample's representation against stored representation in memory. In other words, weights of the base learner would compute based on the similar support set information in memory. Finally, the base learner makes predictions based on the computed weights and seen query points. \cite{bertinetto2016learning} targeted the one-shot classification problem and proposed non-iterative feed-forward neural nets (called learnet) to predict the parameters of a task-specific predictor (named pupil network). While their model architect may be reminiscent of Siamese networks, the meta-learner (i.e., learnet) is dynamic and changes regarding the output of mappings. \cite{li2019lgm} proposed a novel model called LGM-Net consisting of two main components: (\romannum{1}) TargetNet, which operates in task-specific environment; and (\romannum{2}) MetaNet, which is responsible for generating parameters for TargetNet. This MetaNet is different from the Meta-Net model presented by \cite{munkhdalai2017meta} and consists of two modules: (\romannum{1}) task context encoder and (\romannum{2}) weight generator. The task context encoder is responsible for learning representations, while the weight generator is designed to compute the conditional distributions of weights. The authors used matching networks as TargetNet which use generated weights to compute matching probability score as well as classification loss. Another novelty of this study is the introduction of intertask normalization, a strategy that allows similar tasks in a batch of tasks to interact with each other. \par

\cite{gidaris2018dynamic} proposed a dynamic algorithm by introducing an attention-based classification weight generator and a cosine-similarity based ConvNet recognition model. The former is responsible for producing weights vectors for unseen classes by seeing a few examples from those classes. The authors argued that it is not feasible for a ConvNet model with a typical linear layer classifier to handle classification weights of both base and novel classes. They address this issue by proposing the second component that provides prediction by computing the cosine similarity between the generated weights and samples' representations. The combination of these two components allowed the authors to achieve fast adaptation for novel classes without forgetting the base classes. Continuing the previous work and with the same belief of developing the meta-learner without forgetting base classes, \cite{gidaris2019generating} proposed a denoising auto-encoder network (DAE) in the form of graph neural network (GNN), which is responsible for regularizing the weight generating meta-model. Basically, DAE takes contaminated classification weights with Gaussian noise and generates classification weights based on the conditional distribution of seen data from both novel and base classes. Results showed that this model outperforms their previous model in \citep{gidaris2018dynamic}. \par

\begin{figure}[!h]
    \centering
    \includegraphics[width=0.55\linewidth]{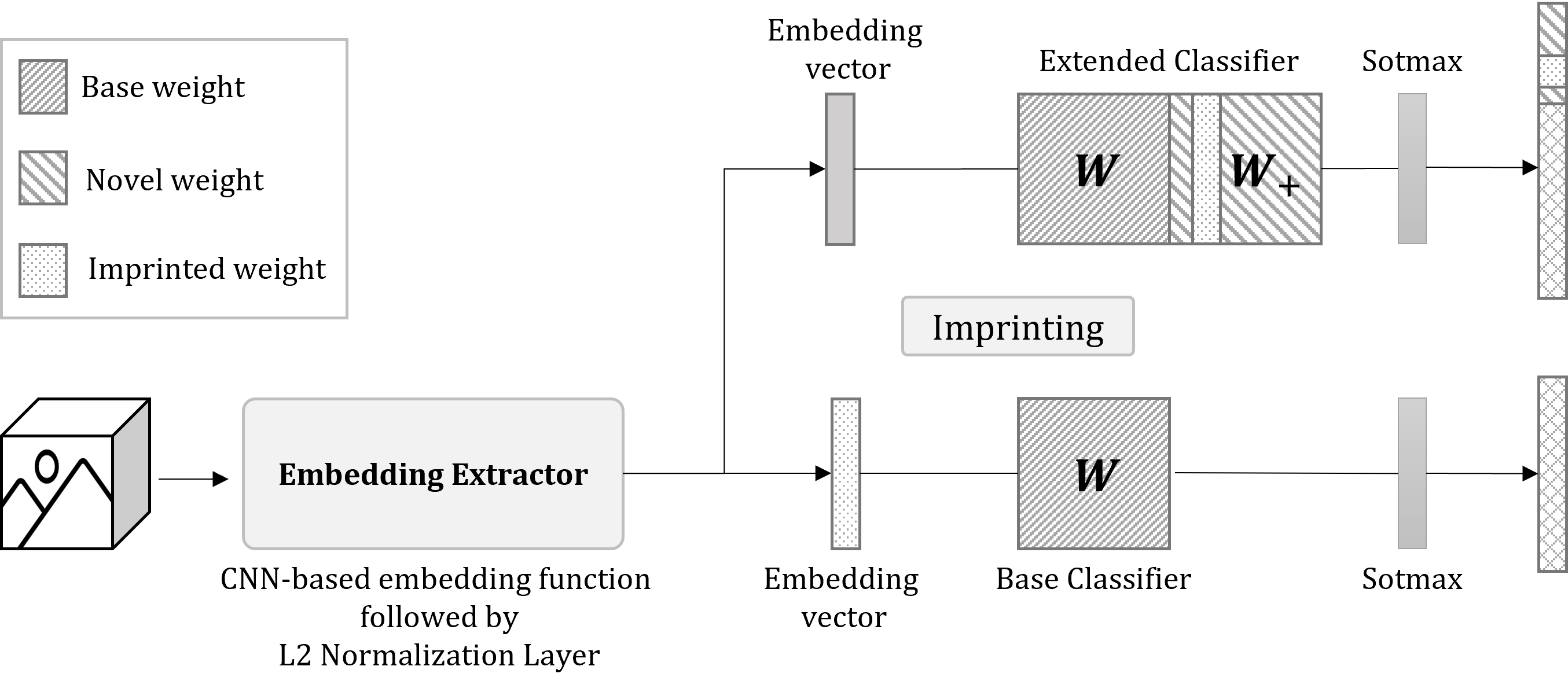}
    \caption{An overview of the weight imprinting \citep{qi2018low} process. Feature vectors of latent space are utilized as imprinted weights in a fully connected layer of the extended classifier which is used for the prediction of new query point. Source: \cite{qi2018low}}
    \label{fig_weight_imprinting}
\end{figure}

\cite{qi2018low} argued that embedding vectors are comparable with the weights of the last linear layer of the ConvNet classifier. Accordingly, instead of using two different components of meta-learner and base learner, they proposed a new process called weight imprinting that basically sets final layer weights based on normalized embeddings of the novel class's training samples. Figure \ref{fig_weight_imprinting} shows the overall flow of their model, in which an embedding function first trains on base classes then tunes on novel classes where imprint weights are used in the final layers. With a similar belief, \cite{ren2019incremental} addressed the issue of forgetting the base classes by proposing incremental learning. Formally, the classifier trains on base classes and learns a set of embeddings. Then. the meta-learner trains on novel classes. This study also introduced an additional regularizer that is optimized by combining both novel and base classes and is computed by an attention attractor network, which solves the catastrophic forgetting problem.  \par

Until now, the reviewed studies in this section focused on generating parameters for a classifier network. \cite{wang2019tafe} proposed a new model called task-aware feature embeddings for low-shot learning (TAFE-Net) that concentrates on generating parameters of feature layers to construct feature embeddings tuned for a given task. TAFE-Net consists of two components: (\romannum{1}) meta-learner that is designed to generate parameters for feature layers in the prediction network; and (\romannum{2}) prediction network, which generates a binary label that indicates that the input image $x$ is compliant to task description $T$. The authors used well-known pre-trained deep neural networks, such as ResNet, to get the features of input images. These generated features feed to the dynamic feature layers. The output of this layer is called task-aware feature embedding since feature embedding of each image may vary depending on the task description. Similarly, \cite{sun2019meta} introduced the meta-transfer learner (MTL) model, focusing on developing task-specific feature extractors by leveraging both transfer learning and meta-learning. Conceptually, they trained a deep neural network on a large dataset, such as miniImageNet dataset, and froze the feature extractor by scaling and shifting operations on pre-trained feature embeddings. This work also introduced a novel hard task meta-batch process, which increases the meta-learner focus on hard tasks through sampling cases that the classifier failed. 

\subsubsection{\textbf{Learning the optimizer}} 
\begin{figure}[!h]
    \centering
    \includegraphics[width=0.75\linewidth]{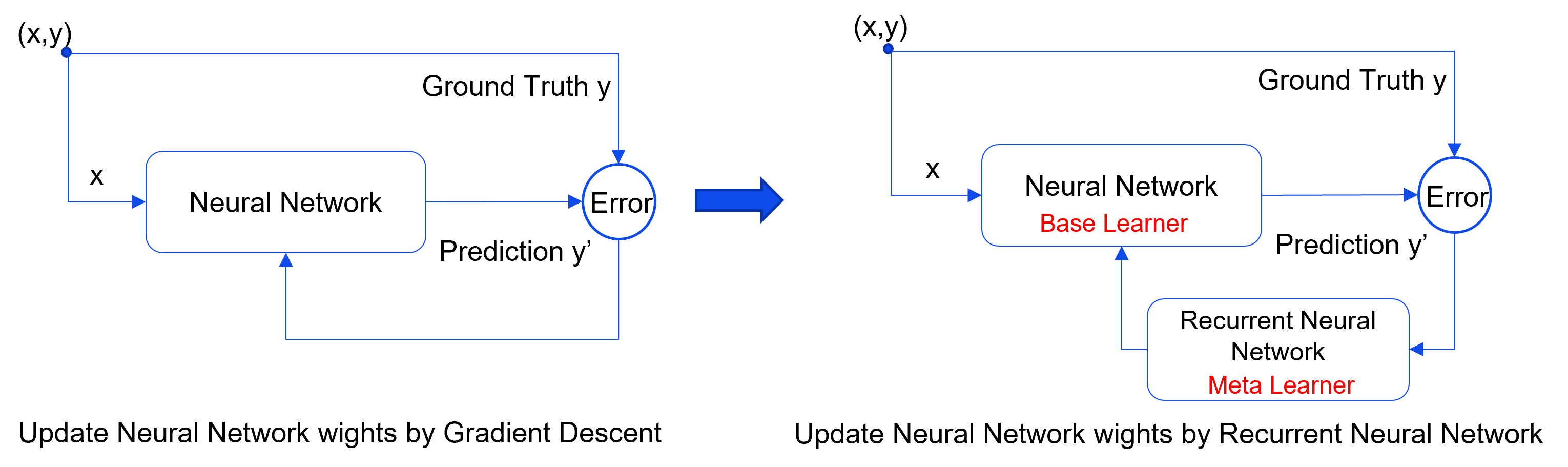}
    \caption{Learning the optimizer conceptual scheme.}
    \label{fig_learning_an_optimizer}
\end{figure}

\hfill\\
A standard neural network trains by calculating loss and using gradient descent to minimize the loss. The key idea of learning the optimizer is to learn gradient descent by replacing it with the recurrent neural network (RNN). Figure \ref{fig_learning_an_optimizer} shows the high-level mechanics of learning the optimizer algorithms. The base learner's parameters are updated through an RNN network (meta-learner). Since RNN optimizes itself through gradient descent, these models are also known as learn gradient descent by gradient descent.  \cite{ravi2016optimization} proposed an LSTM-based meta-learner to learn the optimization process used to train task-specific learner.

Equation \ref{eq_sgd_update_rule} denotes the standard gradient descent update rule:

\begin{equation}
    \theta_t = \theta_{t-1} - \alpha_t\nabla_{\theta_{t-1}}\mathfrak{L_{T_j}}(\theta_{t-1})
    \label{eq_sgd_update_rule}
\end{equation}

\noindent where $\theta_{t-1}$ is the parameter of the learner at $t-1$ step; $\alpha_t$ is the learning rate; and $\nabla_{\theta_{t-1}}\mathfrak{L_{T_j}}$ is the gradient of loss for task $T$ parameterized with $\theta$ at step $t-1$. The authors argued that Eq. \ref{eq_sgd_update_rule} resembles the update cell in LSTM and replaced term $- \alpha_t\nabla_{\theta_{t}}\mathfrak{L_{T_j}}(\theta_{t})$ by LSTM network denoted by $g$ and parameterized with $\varphi$. Thus, Eq. \ref{eq_sgd_update_rule} becomes Eq. \ref{eq_lstm_meta_sgd_update_rule}:

\begin{equation}
    \theta_{t+1} = \theta_t = g_{\varphi}(\nabla_{\theta_{t}}\mathfrak{L_{T_j}}(\theta_{t}))
    \label{eq_lstm_meta_sgd_update_rule}
\end{equation}

\noindent With this new update rule (Eq. \ref{eq_lstm_meta_sgd_update_rule}), the meta-learner (i,e,. LSTM) is able to learn the learning rate $\theta$, which leads to task adaptation quickly. Meta stochastic gradient descent (Meta-SGD) is another learning the optimizer algorithm proposed by \cite{li2017meta}, which is able to learn the learning rate on top of learning initialization too. Recall the standard gradient descent update rule (Eq. \ref{eq_sgd_update_rule}), Meta-SGD proposed a new update rule as Eq. \ref{eq_meta_sgd_update_rule}:

\begin{equation}
    {\theta_t}' = \theta_{t-1} - \alpha \odot \nabla_{\theta_{t-1}}\mathfrak{L_{T_j}}(\theta_{t-1})
    \label{eq_meta_sgd_update_rule}
\end{equation}

\noindent where $\alpha$ is a vector instead of a scalar with the same shape as $\theta$. The operator $\odot$ denotes the element-wise product. In an inner loop, task-specific parameters are updated, and in an outer loop, meta-related optimization is performed on both $\theta$ and $\alpha$ by Eq.\ref{eq_meta_sgd_outer_theta_update} and \ref{eq_meta_sgd_outer_alpha_update}.

\begin{equation}
    \theta_t = \theta_{t-1} - \beta \nabla_{\theta,_{t-1}} \sum_{T_i\sim p(T)} \mathfrak{L_{T_j}}(f({\theta_{i,t-1}}'))
    \label{eq_meta_sgd_outer_theta_update}
\end{equation}

\begin{equation}
    \alpha_t = \alpha_{t-1} - \beta \nabla_{\alpha,_{t-1}}\sum_{T_i\sim p(T)} \mathfrak{L_{T_j}}(f({\theta_{i,t-1}}'))
    \label{eq_meta_sgd_outer_alpha_update}
\end{equation}

Another learning the optimizer algorithm called Reptile proposed by \cite{nichol2018first} tries to generate a task-specific learner close to optimal values. In this regard, the Reptile concept is similar to MAML but quite simpler. Figure \ref{fig_Reptile} shows the mechanism of Reptile. Assume two tasks, $T_1$ and $T_2$, sampled from task distribution $p(T_i)$ randomly, where each task is initialized randomly with parameter $\theta$. For each task, $T_1$ and $T_2$, optimal parameters ${\theta_1}'$ and ${\theta_2}'$ will be obtained through gradient descent after $n$ steps. Then, a random parameter $\theta$ will shift in the direction closer to both obtained parameters, ${\theta_1}'$ and ${\theta_2}'$, by minimizing the Euclidean distance of parameter $\theta$ and task-specific optimal parameters ${\theta_1}'$ and ${\theta_2}'$, as shown in Eq. \ref{eq_reptile_E_euclidean_dist_params}:

\begin{figure}[!h]
    \centering
    \includegraphics[width=0.4\linewidth, height=2cm]{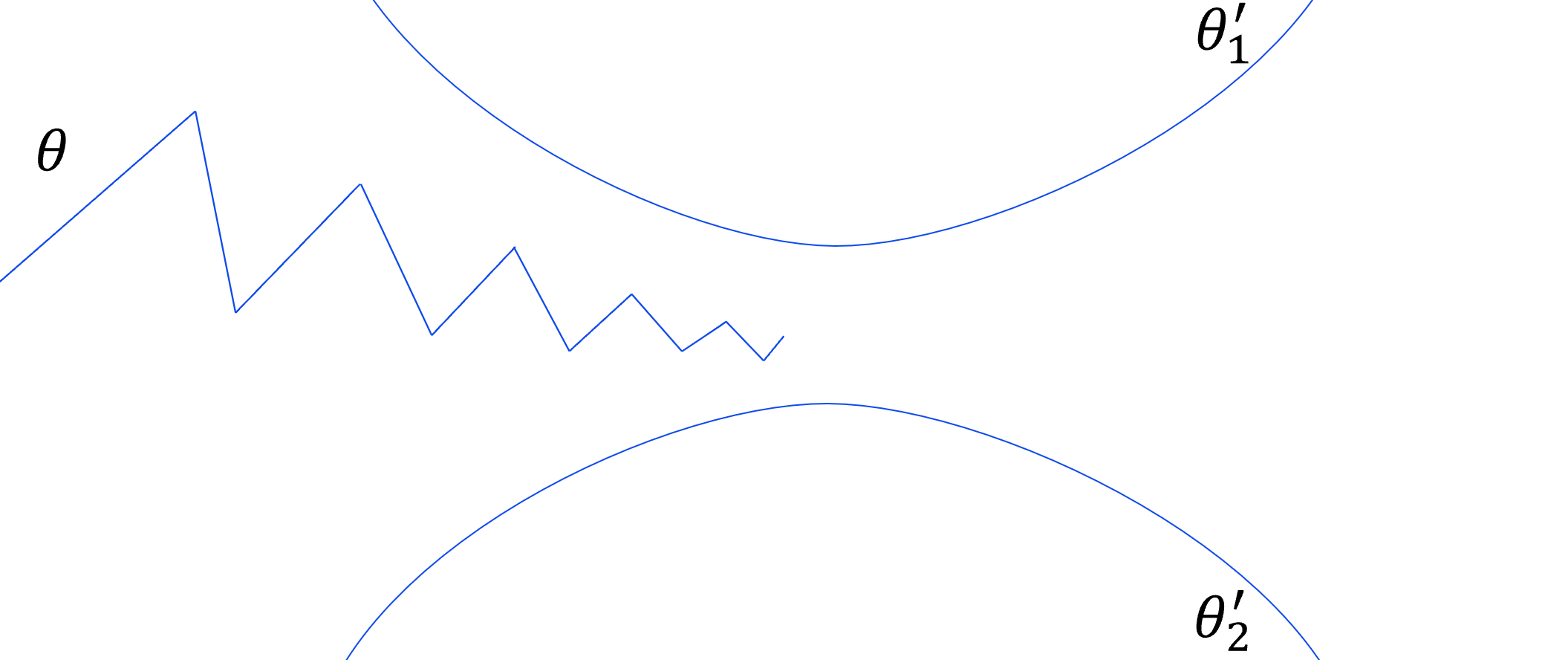}
    \caption{Reptile scheme shows the movement of the random parameter $\theta$ towards tasks optimal parameters.}
    \label{fig_Reptile}
\end{figure}

\begin{equation}
    minimize \: E[\frac{1}{2}D(\theta, {\theta_i}')]
    \label{eq_reptile_E_euclidean_dist_params}
\end{equation}

\noindent Finally, gradient descent is performed on Eq. \ref{eq_reptile_E_euclidean_dist_params} to obtain the following update rule (Eq. \ref{eq_reptile_final_update_rule}): 
 
\begin{equation}
    \theta = \theta - \epsilon ({\theta_i}'-\theta)
    \label{eq_reptile_final_update_rule}
\end{equation}

\noindent where $ \epsilon$ is a step size parameter. \cite{bertinetto2018meta} introduced a new idea about the existing learning the optimizer methods, that is using a non-deep machine learning algorithm to adapt new tasks from few samples. Let's consider a pre-trained network (typically, a CNN-based network) $\phi (x):\mathbb{R}^{m}\rightarrow \mathbb{R}^{e}$
(where $\mathbb{R}^{m}$ and $\mathbb{R}^{e}$ denote the input and embedding space, respectively); and episode-specific predictor $f(\phi (x), w_{\varepsilon }):\mathbb{R}^{e}\times \mathbb{R}^{p} \rightarrow \mathbb{R}^{o}$ (where $\mathbb{R}^{p}$ and $\mathbb{R}^{o}$ refer to the predictor parameters and output spaces, respectively; and $ w_{\varepsilon }\in \mathbb{R}^{p}$ is a set of task-specific predictor parameters). Given the support set of episode, $\varepsilon$, $z_{\varepsilon } = \left \{ (x_i, y_i) \sim \varepsilon  \right \}$ and query set of episode  $\varepsilon$, ${z}'_{\varepsilon } = \left \{ ({x}'_i, {y}'_i) \sim \varepsilon  \right \}$,  $f$ can be considered as a final layer of embedding function denoted by $f(\phi (x)) = \phi (x)W, (w_{\varepsilon }\sim W)$. The authors argued that  $f$ can be trained to obtain parameter $w_{\varepsilon }$ through a learning algorithm $\Lambda $, which can be considered as a typical machine learning algorithm applied in an episodically training fashion and incorporating meta-parameters $w$ and $\rho $. These meta-parameters allow knowledge transfer between episodes. Accordingly, the authors proposed two algorithms, namely Ridge Regression Differentiable Discriminator (R2-D2) and Logistic Regression Differentiable Discriminator (LR-D2). In R2-D2, the optimal $W$ is obtained by Eq.\ref{eq_r2_d2_first_equation}:
\begin{equation}
    \begin{aligned}
    \Lambda(z) &= argmin \left \| XW-Y \right \|^2 + \lambda \left \| W \right \|^2 \\ 
               &= (X^TX+\lambda I)^{-1}X^TY
    \end{aligned}
    \label{eq_r2_d2_first_equation}
\end{equation}
where $X$, $Y$, and $\lambda$ are embeddings, outputs, and regularization terms, respectively. Considering the high dimensionality of embeddings,  $X^TX\in \mathbb{R}^{e\times e}$ can be computationally expensive since the matrix $X^TX$ grows quadratically. The authors addressed this issue by using the Woodbury formula, as given in  Eq.\ref{eq_r2_d2_woodbury}:
\begin{equation}
    W = \Lambda(z) = X^T(XX^T+ \lambda I)^{-1}Y
    \label{eq_r2_d2_woodbury}
\end{equation}
With small number of $n$ in the $n$-shot problem, the computation cost of Eq.\ref{eq_r2_d2_woodbury} is generally small. Although Eq.\ref{eq_r2_d2_first_equation} originally works for regression tasks, the authors modified it to Eq.\ref{eq_r2_d2_classification} for classification tasks:

\begin{equation}
    \hat{Y} = \alpha {X}'W+\beta 
    \label{eq_r2_d2_classification}
\end{equation}
The meta-learner learns $\alpha$, $\beta$, and $\lambda$ (hyper-parameters of base learner) along with CNN parameters $W$. The authors asserted that other similar solvers can be employed for differentiable operations. Thus, they proposed the LR-D2 method, which is basically logistic regression that uses Newton's method as an iterative solver. LR-D2, unlike R2-D2, is designed to apply to classification tasks directly. Considering inputs $X\in \mathbb{R}^{n\times e}$ and binary label $y\in \left \{ -1,+1 \right \}^n$, parameters $w_i\in \mathbb{R}^e$ are updated by Eq.\ref{eq_l2_d2_first_eq}:

\begin{equation}
     w_i = (X^T diag(s_i)X + \lambda I)^{-1}X^T diag (s_i)z_i
    \label{eq_l2_d2_first_eq}
\end{equation}
where $w_i$ denotes the parameters at iteration $i$; and I stands for the identity matrix. Moreover, $s_i = \mu_i(1-\mu_i)$, $z_i = w_{i-1}^TX+(y-\mu_i)/s_i$ and $\mu_i = \sigma (w_{i-1}^TX)$, where $\sigma$ is the sigmoid function. Similarly, the authors applied Woodbury's identity on $X^T diag(s_i)X$ and modified Eq.\ref{eq_l2_d2_first_eq} to Eq.\ref{eq_l2_d2_woodbury}:

\begin{equation}
    w_i = X^T(XX^T + \lambda diag(s_i)^{-1})^{-1}z_i
    \label{eq_l2_d2_woodbury}
\end{equation}
With the same idea of using a machine learning classifier as a base learner, \cite{lee2019meta} proposed a new method called MetaOptNet. In other words, a support vector machine (SVM) is applied as a base learner incorporated with a differentiable quadratic programming (QP) solver to learn parameters of the embedding function $f_\phi$. This method showed to outperform both R2-D2 and LR-D2.

\section{Overview, Challenges and Future Directions} \label{section_challenges}
\begin{figure}[!h]
    \centering
    \includegraphics[width=0.8\linewidth]{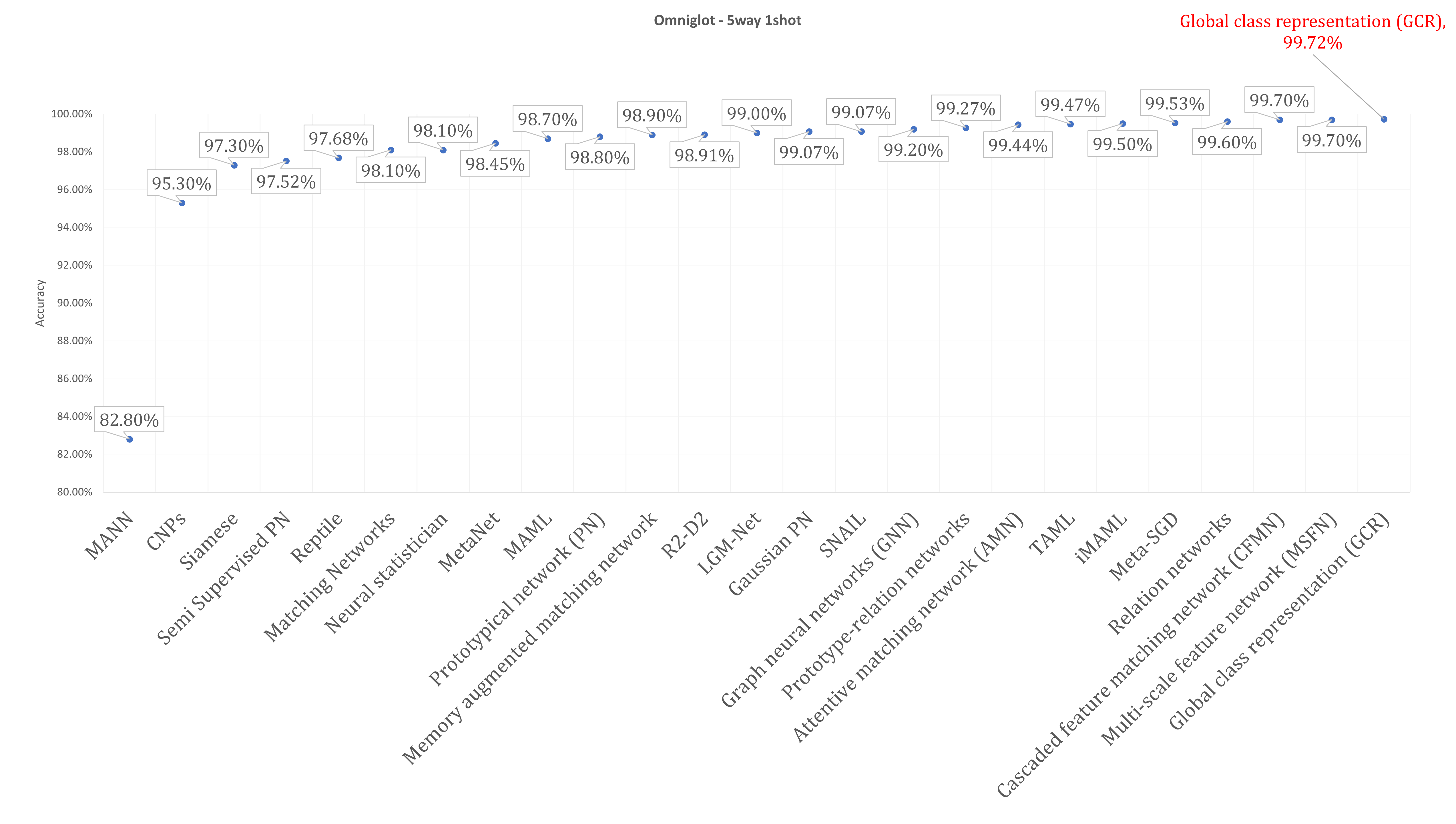}
    \caption{5way-1shot Omniglot Accuracy results}
    \label{fig_5way-1shot-omniglot}
\end{figure}

\begin{figure}[!h]
    \centering
    \includegraphics[width=0.8\linewidth]{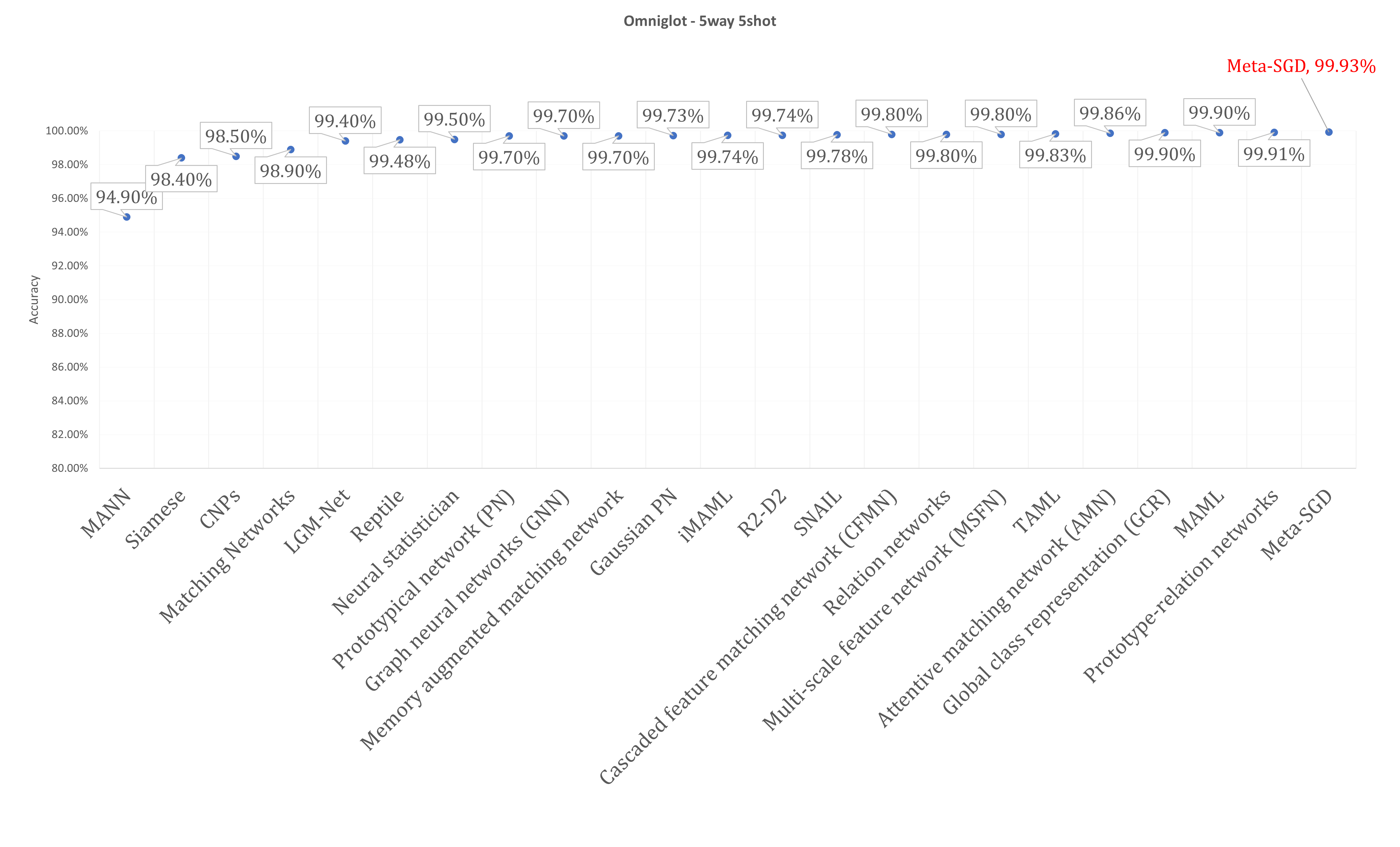}
    \caption{5way-5shot Omniglot Accuracy results}
    \label{fig_5way-5shot-omniglot}
\end{figure}

\begin{figure}[!h]
    \centering
    \includegraphics[width=0.8\linewidth]{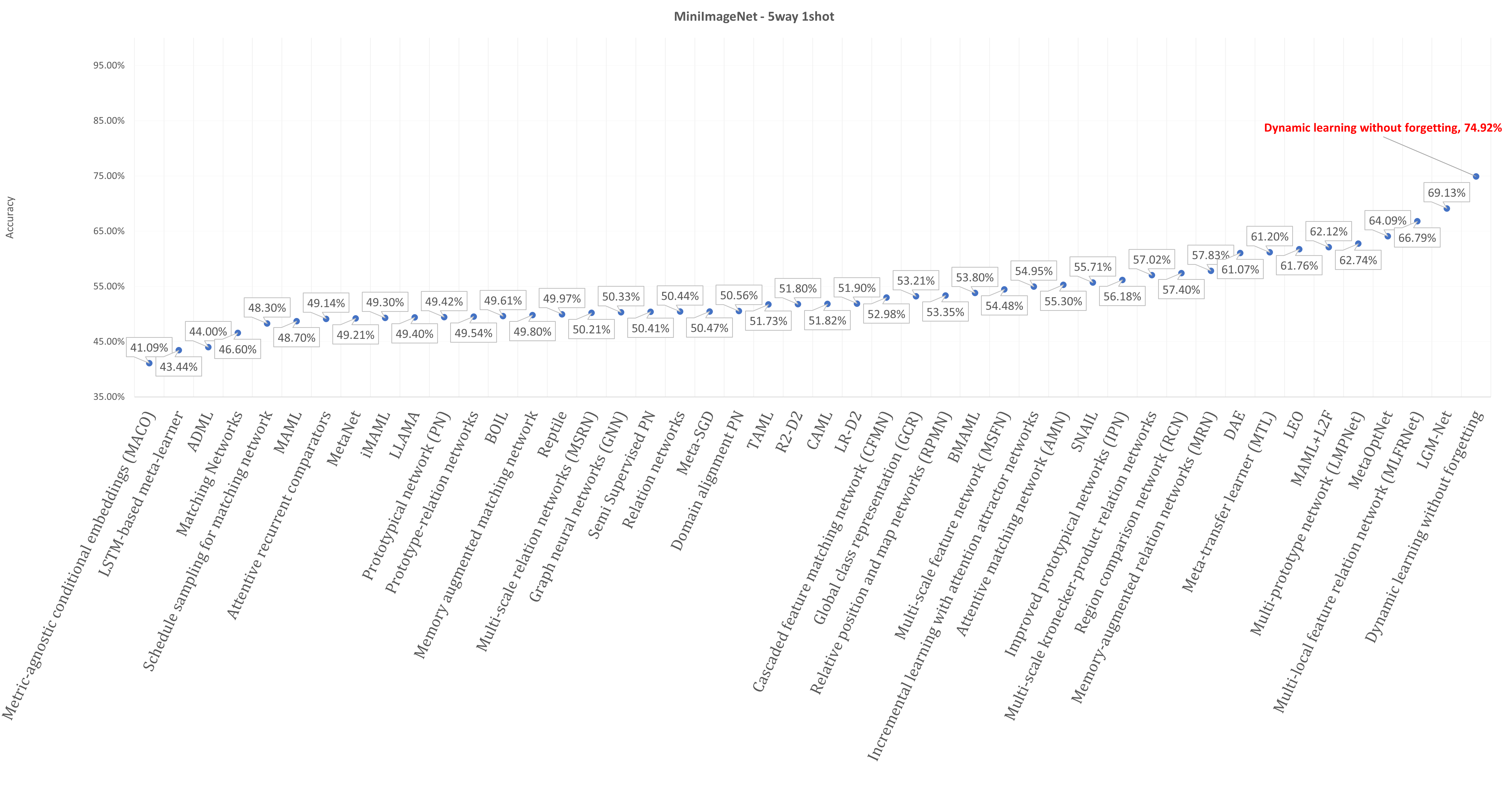}
    \caption{5way-1shot MiniImageNet Accuracy results}
    \label{fig_5way-1shot-MiniImageNet}
\end{figure}

\begin{figure}[!h]
    \centering
    \includegraphics[width=0.8\linewidth]{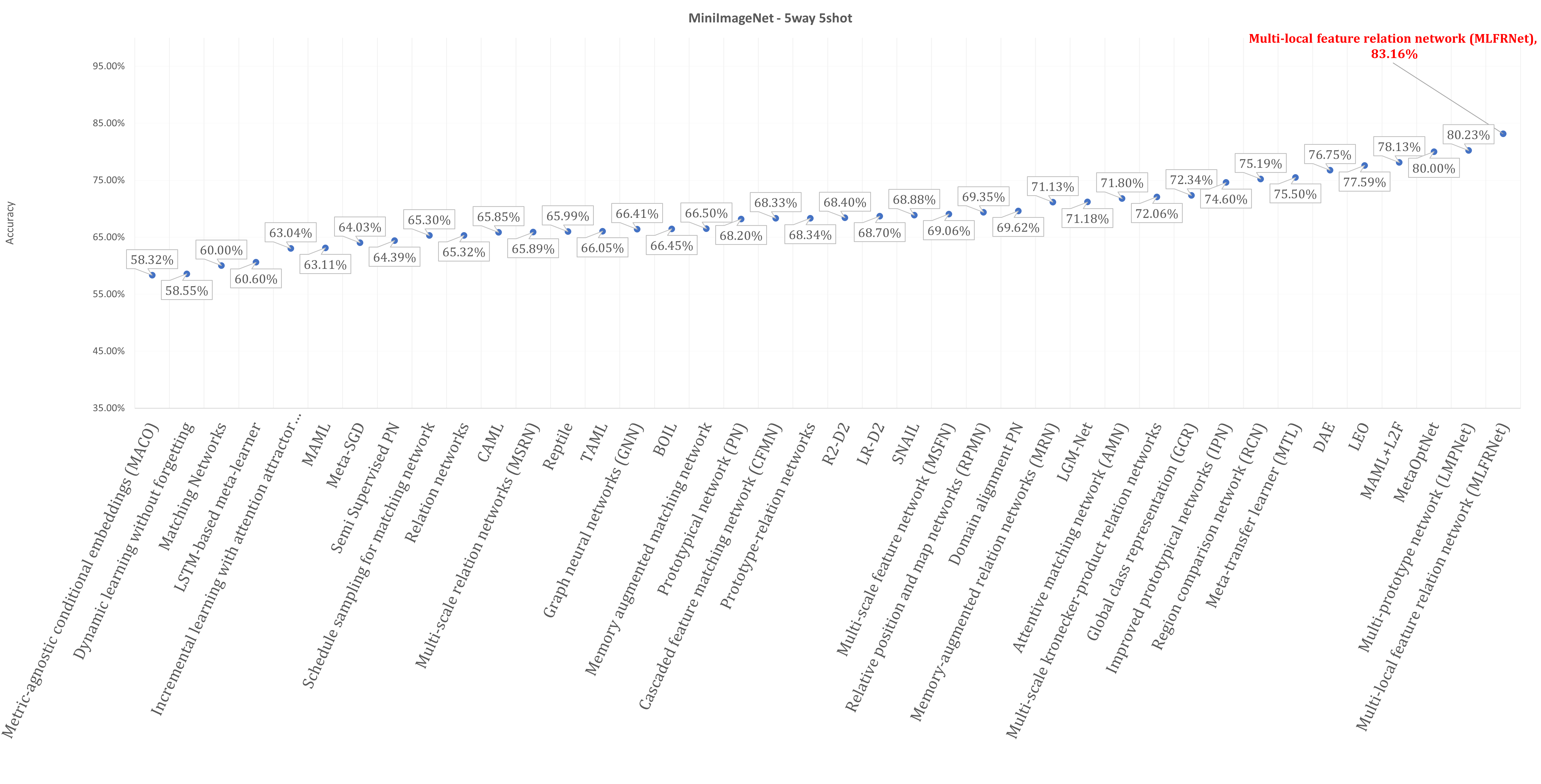}
    \caption{5way-5shot MiniImageNet Accuracy results}
    \label{fig_5way-5shot-MiniImageNet}
\end{figure}

Along with the considerable success of deep learning, its impotence in domain generalization and avaricious craving for a large dataset for training have increased the demand for few-shot meta-learning algorithms. Despite the fact that in recent years, meta-learning has attracted remarkable interest from researchers, meta-learning is still in its infancy. 
Although the Siamese network was originally proposed in 1993, the paper "Siamese neural networks for one-shot image recognition" in 2015 \cite{koch2015siamese} triggered a new era of meta-learning and metric-based algorithms. While Siamese networks are limited to binary classification tasks, other metric-based methods resolve this limitation by training on tasks in an episodic training manner. Prototypical networks bring the idea of using class prototypes to compare every new query point instead of individual comparisons in Siamese networks. This idea was the starting point of many metric-based methods. Relation networks employ a recurrent neural network instead of a fixed similarity function, leading to a dynamic nature of learning domain-related similarity functions. A general view on recent metric-based meta-learners demonstrates that the concentration on inter-class and intra-class relationships and a proposed weighted prototype has been increased. Compared to other non-metrid based models, metric-based methods are simpler but more innovative in proposing novel similarity measures. Given the pair-wise comparisons – whether a comparison with class prototypes or direct comparison with support data points – metric-based methods can be computationally expensive, especially with large tasks. The next group, memory-based meta-learning models, have received less attention than other groups,but show strong performance. The key advantage of memory-based models is their robustness to forget base classes as the fruit of external/internal memory. Finally, the last large group, the learning-based methods, comprises the major area of research. The LSTM-based meta-learner was the first to materialize the idea of learning gradient descent by gradient descent. It will not be an overstatement if MAML is considered the most inspiring method and commencement mark of learning the initialization methods. While the learning-based methods outperform other methods on a broader range of tasks \cite{finn2018meta}, they can be computationally expensive since they optimize a base learner for every task \cite{finn2018meta}. Table\ref{tab_benchmark_result} compares the performance of the discussed methods on benchmark datasets extracted from the papers. Figures \ref{fig_5way-1shot-omniglot}, \ref{fig_5way-5shot-omniglot}, \ref{fig_5way-1shot-MiniImageNet} and \ref{fig_5way-5shot-MiniImageNet} summarize the results for the omniglot and MiniImageNet 5-way problems. The first impression from the figures is that the 5-shot setting leads to better performance. Regarding the omniglot dataset, Meta-SGD (99.91\%), Prototype-relation networks (99.90\%), MAML (99.90\%), and Global class representation (99.86\%) achieve the highest accuracy in 5-way 5-shot problem. It can be concluded that when the tasks are closely related, most proposed methods perform well and closely regardless of their mechanics. However, in more complex datasets where tasks are more distant like miniImageNet, the circumstances change. The multi-local feature relation network (MLFRNet) and multi-prototype network (LMPNet) showed the best performance of metric-based methods on 5-way 5-shot miniImageNet problem with 80.23\% and 80.00\% accuracy, respectively. This result confirms the above statement that metric-based approaches, despite having less computational complexity, have shown to be very powerful since they have the ability to pay attention to local features and minimize the intra-classes' difference while maximizing the inter-classes' difference. Moreover, further improvement and development of these methods are still needed to achieve higher accuracy, yet it remains unclear whether the current methods are confident in their decisions or not. 


{\footnotesize
\begin{landscape}
\setlength\tabcolsep{0pt}
\begin{longtable}[c]{lccccccccc}
\caption{Accuracy results on benchmarks datasets reported in original papers, \romannum{1}: Metric-based;  \romannum{2}: Memory-based;  \romannum{3}: Learning the initialization;  \romannum{4}: Learning the parameter;  \romannum{5}: Learning the optimizer; The ± indicates the 95\% confidence interval. “-” indicates not reported. The MANN\cite{santoro2016meta} performance is extracted from \cite{garnelo2018conditional}.}
\label{tab_benchmark_result}\\
\hline
\multicolumn{1}{c}{\multirow{3}{*}{Model}} &
  \multirow{3}{*}{Category} &
  \multicolumn{4}{c}{Omniglot} &
  \multicolumn{2}{c}{MiniImaget} &
  \multicolumn{2}{c}{CUB-200-2011} \\ \cline{3-10} 
\multicolumn{1}{c}{} &
   &
  \multicolumn{2}{c}{5-way} &
  \multicolumn{2}{c}{20-way} &
  \multicolumn{2}{c}{5-way} &
  \multicolumn{2}{c}{5-way} \\ \cline{3-10} 
\multicolumn{1}{c}{} &
   &
  1-shot &
  5-shot &
  1-shot &
  5-shot &
  1-shot &
  5-shot &
  1-shot &
  5-shot \\ \hline
\endhead
\hline
\endfoot
\endlastfoot
Siamese\cite{koch2015siamese} &
  \romannum{1} &
  97.3\% &
  98.4\% &
  88.2\% &
  97\% &
  - &
  - &
  - &
  - \\
Prototypical network (PN)\cite{snell2017prototypical} &
  \romannum{1} &
  98.8\% &
  99.7\% &
  96\% &
  98.9\% &
  49.42 ± 0.78\% &
  68.20 ± 0.66\% &
  - &
  - \\
Semi Supervised PN\cite{ren2018meta} &
  \romannum{1} &
  97.52 ± 0.07\% &
  - &
  - &
  - &
  50.41 ± 0.31 &
  64.39 ± 0.24 &
  - &
  - \\
Gaussian PN\cite{fort2017gaussian} &
  \romannum{1} &
  99.07 ± 0.03\% &
  99.73 ± 0.02\% &
  96.94 ± 0.14\% &
  99.29 ± 0.04\% &
  - &
  - &
  - &
  - \\
Transferable prototypical network   (TPN)\cite{pan2019transferrable} &
  \romannum{1} &
  - &
  - &
  - &
  - &
  - &
  - &
  - &
  - \\
Domain alignment PN\cite{lu2018boosting} &
  \romannum{1} &
  - &
  - &
  - &
  - &
  50.56 ± 0.85\% &
  69.62 ± 0.76\% &
  - &
  - \\
Improved prototypical networks (IPN)\cite{ji2020improved} &
  \romannum{1} &
  - &
  - &
  - &
  - &
  56.18 ± 0.85\% &
  74.60 ± 0.83\% &
  73.25\% &
  86.81\% \\
Label enhanced prototypical network (LPN)\cite{liu2022label} &
  \romannum{1} &
  - &
  - &
  - &
  - &
  - &
  - &
  - &
  - \\
Multi-prototypes PN\cite{deuschel2021multi} &
  \romannum{1} &
  - &
  - &
  - &
  - &
  - &
  - &
  - &
  - \\
Dummy prototypical networks\cite{kim2022dummy} &
  \romannum{1} &
  - &
  - &
  - &
  - &
  - &
  - &
  - &
  - \\
Multi-prototype network (LMPNet)\cite{huang2021local} &
  \romannum{1} &
  - &
  - &
  - &
  - &
  62.74 ± 0.11\% &
  80.23 ± 0.52\% &
  65.59 ± 0.13\% &
  68.19 ± 0.23\% \\
Multi-modal prototypical network.\cite{pahde2021multimodal} &
  \romannum{1} &
  - &
  - &
  - &
  - &
  - &
  - &
  75.01 ± 0.81\% &
  85.30 ± 0.54\% \\
Matching Networks\cite{vinyals2016matching} &
  \romannum{1} &
  98.1\% &
  98.9\% &
  93.8\% &
  98.7\% &
  46.6\% &
  60\% &
  - &
  - \\
Cascaded feature matching network   (CFMN)\cite{chen2021learning} &
  \romannum{1} &
  99.7 ± 0.2\% &
  99.8 ± 0.1\% &
  98.0 ± 0.2\% &
  99.2 ± 0.1\% &
  52.98 ± 0.84\% &
  68.33 ± 0.70\% &
  - &
  - \\
Attentive matching network (AMN)\cite{mai2019attentive} &
  \romannum{1} &
  99.44 ±0.09\% &
  99.86 ±0.06\% &
  98.06 ±0.11\% &
  99.50 ±0.07\% &
  55.30 ±0.89\% &
  71.80 ±0.78\% &
  - &
  - \\
Schedule sampling for matching   network\cite{zhang2019scheduled} &
  \romannum{1} &
  - &
  - &
  - &
  - &
  48.3\% &
  65.3\% &
  - &
  - \\
Relation networks\cite{sung2018learning} &
  \romannum{1} &
  99.6 ± 0.2\% &
  99.8 ± 0.1\% &
  97.6 ± 0.2\% &
  99.1 ± 0.1\% &
  50.44 ± 0.82\% &
  65.32 ± 0.70\% &
  - &
  - \\
Memory-augmented relation networks (MRN)\cite{he2020memory} &
  \romannum{1} &
  - &
  - &
  - &
  - &
  57.83 ± 0.69\% &
  71.13 ± 0.50\% &
  - &
  - \\
Prototype-relation networks\cite{liu2020meta} &
  \romannum{1} &
  99.27 ± 0.23\% &
  99.91 ± 0.06\% &
  95.97 ± 0.56\% &
  99.32 ± 0.13\% &
  49.54 ± 0.09\% &
  68.34 ± 0.06\% &
  - &
  - \\
Multi-scale relation networks (MSRN)\cite{ding2019multi} &
  \romannum{1} &
  - &
  - &
  - &
  - &
  50.21 ± 1.08\% &
  65.89 ± 0.32\% &
  - &
  - \\
Multi-local feature relation network   (MLFRNet)\cite{ren2022multi} &
  \romannum{1} &
  - &
  - &
  - &
  - &
  66.79 ± 0.47\% &
  83.16 ± 0.39\% &
  - &
  - \\
Graph neural networks (GNN)\cite{garcia2017few} &
  \romannum{1} &
  99.2\% &
  99.7\% &
  97.4\% &
  99\% &
  50.33 ±0.36\% &
  66.41 ±0.63\% &
  - &
  - \\
Global class representation (GCR)\cite{li2019few} &
  \romannum{1} &
  99.72 ±0.06\% &
  99.90 ±0.10\% &
  99.63 ±0.09\% &
  99.32 ±0.04\% &
  53.21 ± 0.40\% &
  72.34 ± 0.32\% &
  - &
  - \\
Multi-scale feature network (MSFN)\cite{han2020multi} &
  \romannum{1} &
  99.7\% &
  99.8\% &
  98.1\% &
  99.2\% &
  54.48 ± 1.23\% &
  69.06 ± 0.69\% &
  62.4\% &
  79.14\% \\
Attentive recurrent comparators\cite{shyam2017attentive} &
  \romannum{1} &
  - &
  - &
  - &
  - &
  49.14\% &
  - &
  - &
  - \\
Relative position and map networks   (RPMN)\cite{xue2020relative} &
  \romannum{1} &
  - &
  - &
  - &
  - &
  53.35\% &
  69.35\% &
  - &
  - \\
Region comparison network (RCN)\cite{xue2020region} &
  \romannum{1} &
  - &
  - &
  - &
  - &
  57.40 ± 0.86\% &
  75.19 ± 0.64\% &
  74.65 ± 0.86\% &
  88.81 ± 0.57\% \\
Multi‑scale kronecker‑product relation   networks\cite{abdelaziz2022multi} &
  \romannum{1} &
  - &
  - &
  - &
  - &
  57.02 ± 0.88\% &
  72.06 ± 0.68\% &
  69.49 ± 0.95\% &
  82.94 ± 0.65\% \\
Metric-agnostic conditional embeddings   (MACO)\cite{hilliard2018few} &
  \romannum{1} &
  - &
  - &
  - &
  - &
  41.09\% &
  58.32\% &
  60.76\% &
  74.96\% \\
MANN\cite{santoro2016meta} &
  \romannum{2} &
  82.8\% &
  94.9\% &
  - &
  - &
  - &
  - &
  - &
  - \\
SNAIL\cite{mishra2017simple} &
  \romannum{2} &
  99.07 ± 0.16\% &
  99.78 ± 0.09\% &
  97.64 ± 0.30\% &
  99.36 ± 0.18\% &
  55.71 ± 0.99\% &
  68.88 ± 0.92\% &
  - &
  - \\
CNPs\cite{garnelo2018conditional} &
  \romannum{2} &
  95.3\% &
  98.5\% &
  89.9\% &
  96.8\% &
  - &
  - &
  - &
  - \\
Memory augmented matching network\cite{tran2019memory} &
  \romannum{2} &
  98.9\% &
  99.7\% &
  96.3\% &
  98.9\% &
  49.8\% &
  66.5\% &
  - &
  - \\
MAML\cite{finn2017model} &
  \romannum{3} &
  98.7 ± 0.4\% &
  99.9 ± 0.1\% &
  95.8 ± 0.3\% &
  98.9 ± 0.2\% &
  48.70 ± 1.84\% &
  63.11 ± 0.92\% &
  - &
  - \\
PLATIPUS\cite{finn2018probabilistic} &
  \romannum{3} &
  - &
  - &
  - &
  - &
  - &
  - &
  - &
  - \\
BMAML\cite{yoon2018bayesian} &
  \romannum{3} &
  - &
  - &
  - &
  - &
  53.80 ± 1.46\% &
  - &
  - &
  - \\
LEO\cite{rusu2018meta} &
  \romannum{3} &
  - &
  - &
  - &
  - &
  61.76 ± 0.08\% &
  77.59 ± 0.12\% &
  - &
  - \\
CAML\cite{zintgraf2018caml} &
  \romannum{3} &
  - &
  - &
  - &
  - &
  51.82 ± 0.65\% &
  65.85 ± 0.55\% &
  - &
  - \\
ADML\cite{yin2018adversarial} &
  \romannum{3} &
  - &
  - &
  - &
  - &
  44.00 ± 1.83\% &
  - &
  - &
  - \\
TAML\cite{jamal2019task} &
  \romannum{3} &
  99.47 ± 0.25 \% &
  99.83 ± 0.09\% &
  95.62 ± 0.5\% &
  98.64 ± 0.13\% &
  51.73 ± 1.88\% &
  66.05 ± 0.85\% &
  - &
  - \\
Alpha MAML\cite{behl2019alpha} &
  \romannum{3} &
  - &
  - &
  - &
  - &
  - &
  - &
  - &
  - \\
BOIL\cite{oh2020boil} &
  \romannum{3} &
  - &
  - &
  - &
  - &
  49.61 ± 0.16\% &
  66.45 ± 0.37\% &
  - &
  - \\
MAML+L2F\cite{baik2020learning} &
  \romannum{3} &
  - &
  - &
  - &
  - &
  62.12 ± 0.13\% &
  78.13 ± 0.15\% &
  - &
  - \\
iMAML\cite{rajeswaran2019meta} &
  \romannum{3} &
  99.50 ± 0.26\% &
  99.74 ± 0.11\% &
  96.18 ± 0.36\% &
  99.14 ± 0.1\% &
  49.30 ± 1.88 \% &
  - &
  - &
  - \\
LLAMA\cite{grant2018recasting} &
  \romannum{3} &
  - &
  - &
  - &
  - &
  49.40 ± 1.83\% &
  - &
  - &
  - \\
MetaNet\cite{munkhdalai2017meta} &
  \romannum{4} &
  98.45\% &
  - &
  95.92\% &
  - &
  49.21 ± 0.96\% &
  - &
  - &
  - \\
Pupil network\cite{bertinetto2016learning} &
  \romannum{4} &
  - &
  - &
  - &
  - &
  - &
  - &
  - &
  - \\
LGM-Net\cite{li2019lgm} &
  \romannum{4} &
  99\% &
  99.4\% &
  96.5\% &
  98.5\% &
  69.13 ± 0.35\% &
  71.18 ± 0.68\% &
  - &
  - \\
Dynamic learning without forgetting\cite{gidaris2018dynamic} &
  \romannum{4} &
  - &
  - &
  - &
  - &
  74.92 ± 0.36\% &
  58.55 ± 0.50\% &
  - &
  - \\
DAE\cite{gidaris2019generating} &
  \romannum{4} &
  - &
  - &
  - &
  - &
  61.07 ± 0.15\% &
  76.75 ± 0.11\% &
  - &
  - \\
Weight imprinting\cite{qi2018low} &
  \romannum{4} &
  - &
  - &
  - &
  - &
  - &
  - &
  - &
  - \\
Incremental learning with attention attractor   networks\cite{ren2019incremental} &
  \romannum{4} &
  - &
  - &
  - &
  - &
  54.95 ± 0.30\% &
  63.04 ± 0.30\% &
  - &
  - \\
TAFE-Net\cite{wang2019tafe} &
  \romannum{4} &
  - &
  - &
  - &
  - &
  - &
  - &
  - &
  - \\
Meta-transfer learner (MTL)\cite{sun2019meta} &
  \romannum{4} &
  - &
  - &
  - &
  - &
  61.2 ± 1.8\% &
  75.5 ± 0.8\% &
  - &
  - \\
Neural statistician\cite{edwards2016towards} &
  \romannum{4} &
  98.1\% &
  99.5\% &
  93.2\% &
  98.1\% &
  - &
  - &
  - &
  - \\
LSTM-based meta-learner\cite{ravi2016optimization} &
  \romannum{5} &
  - &
  - &
  - &
  - &
  43.44 ± 0.77\% &
  60.60 ± 0.71\% &
  - &
  - \\
Meta-SGD\cite{li2017meta} &
  \romannum{5} &
  99.53 ± 0.26\% &
  99.93 ± 0.09\% &
  95.93 ± 0.38\% &
  98.97 ± 0.19\% &
  50.47 ± 1.87\% &
  64.03 ± 0.94\% &
  - &
  - \\
Reptile\cite{nichol2018first} &
  \romannum{5} &
  97.68 ± 0.04\% &
  99.48 ± 0.06\% &
  89.43 ± 0.14\% &
  97.12 ± 0.32\% &
  49.97 ± 0.32\% &
  65.99 ± 0.58\% &
  - &
  - \\
MetaOptNet\cite{lee2019meta} &
  \romannum{5} &
  - &
  - &
  - &
  - &
  64.09 ± 0.62\% &
  80.00 ± 0.45\% &
  - &
  - \\
R2-D2\cite{bertinetto2018meta} &
  \romannum{5} &
  98.91 ± 0.05\% &
  99.74 ± 0.02\% &
  96.24 ± 0.05\% &
  99.20 ± 0.02\% &
  51.8 ± 0.2\% &
  68.4 ± 0.2\% &
  - &
  - \\
LR-D2\cite{bertinetto2018meta} &
  \romannum{5} &
  - &
  - &
  - &
  - &
  51.9 ± 0.2\% &
  68.7 ± 0.2\% &
  - &
  - \\ \hline
\end{longtable}
\end{landscape}
}

Despite the promising achievements of meta-learning, some challenges remain. These existing issues along with suggested future research directions are presented here.

\begin{itemize}
    \item \textbf{Uncertainty of meta-learners' performance.} 
    As the name of few-shot learning implies, the main challenge is learning from a handful of samples. This issue has been addressed by meta-learning via sharing knowledge or generating knowledge for each unseen task from previously experienced tasks. In this context, experience is interpreted as fitting to data. However, during episodic training, unrelated tasks or even less related tasks may result in divergent optimization directions as well as disparate performance, which eventually may lead to uncertainty in unseen class prediction \cite{jamal2019task}. Therefore, training a meta-learner with a confident performance in novel classes needs to be developed. One suggestion worth exploring is training a meta-learner on more diversified tasks and grouping more similar tasks to train in every episode.  
    \item \textbf{Effective use of episodic training.} 
    Episodic training is prone to catastrophic forgetting, which results in model underfitting in base classes. While several researches have addressed this issue \cite{ji2020improved, gidaris2018dynamic, gidaris2019generating}, enhancing model performance on both base and unseen classes remains an important direction for future work. Combining memory-based methods with metric-based methods may be valuable exploration. 
    \item \textbf{Improving stability.} Although GCR, a metric-based meta-learner, achieved the best performance on the Omniglot dataset, metric-based models cannot compete with other non-metric-based in miniImageNet, a colored dataset that is more complex than Omniglot. In other words, metric-based meta-learning algorithms are sensitive to the dataset. Moreover, it has been shown that other existing meta-leaning algorithms, such as MAML, are not robust to adversarial samples \cite{goldblum2020adversarially}. While ADML investigates the adversarial samples to improve meta-learner performance, more exploration will be valuable. 
    \item \textbf{Training multi-domain, multi-modal and cross-domain meta-learning.} 
    In theory, $D_Base$ and $D_Novel$ can be from separate domains. However, in practice, as the difference between $D_Base$ and $D_Novel$ increases, most meta-learning performance will decline. This can be seen by the poor performance of metric-based methods on the miniImageNet dataset. Additionally, the current methods are trained on a single domain $D_Base$. As a suggestion for future research exploration, meta-learning can be developed on multi-domain and cross-domain performance. While training on a multi-domain $D_Base$ is expected to increase generalization ability, it can be quite challenging due to divergent optimization direction or the possibility of forgetting $D_Base$ tasks. Therefore, using memory-based methods or grouping related tasks based on statistical information is worth exploring. Although \cite{triantafillou2019meta} recently proposed a new benchmark dataset by combining Omniglot and ImageNet, which is useful for multi-domain and cross-domain evaluation, other benchmark datasets from a combination of real-world applications, such as remote sensing scene images and agricultural pest images, and other real-world domains should be investigated. Last, only a few researches have studied  multi-modal domains \cite{pahde2021multimodal, finn2018probabilistic, yu2018beyond}. Therefore, the development of multi-modal benchmark datasets and methods on multi-modal datasets will be valuable to study. Further research on the multi-modal datasets and real-world multi-domain and cross-domains datasets is pertinent to determine if the current methods are successful in learning from a few samples or simply re-using learned features. 
    \item \textbf{Representation learning effects.} 
    Different meta-learning methods use embedding functions with different architectures. While most methods use four-layer convolutional neural networks as a backbone, few studies have applied larger networks, such as resnet, and multi-scale networks as backbone \cite{ding2019multi, han2020multi}. Accordingly, it may be insightful to analyze the effect of embedding functions, like multi-scale networks, or changing the architecture of the backbone or pre-trained neural networks on state-of-the-art methods such as prototypical networks performance as well as computational costs. 
    \item \textbf{Improvement in computational cost.} 
    As mentioned earlier in this section, learning-based methods can be quite computationally expensive in spite of learning from a few samples. Improving computational cost is another important direction for future research.  
\end{itemize}

In conclusion, this study investigated state-of-the-art methods and advances in meta-learning. A clear description of meta-learning is also provided along with a thorough organization of the methods. Whereas deep learning methods focus on data, meta-learning concentrates more on tasks, thus it can be comparatively confusing with other similar approaches such as transfer learning or multi-task learning for researchers who are new to few-shot learning. It is hope that this study and the discussed research challenges and future directions encourage further development in this area.

\bibliographystyle{unsrt}
\bibliography{Refs} 

\end{document}